%% file: main.tex
\pdfoutput=1

\documentclass[11pt]{article}

\usepackage[final]{acl}

\usepackage{times}
\usepackage{latexsym}
\usepackage[T1]{fontenc}

\usepackage[utf8]{inputenc}

\usepackage{microtype}
\usepackage{inconsolata}

\usepackage{graphicx}
\usepackage{enumitem}
\usepackage{kotex}
\usepackage{booktabs}
\usepackage{xcolor}
\usepackage{colortbl}
\usepackage{multirow}
\usepackage{relsize}
\usepackage{tcolorbox}
\usepackage{rotating}

\usepackage{array}
\usepackage{caption}
\usepackage{tabularx}

\newcommand{\original}{\textsc{KMMLU}}

\newcommand{\pro}{\textsc{KMMLU-Pro}}
\newcommand{\redux}{\textsc{KMMLU-Redux}}

\newcommand{\correspondingfootnote}{
    \let\oldthefootnote=\thefootnote
    \renewcommand{\thefootnote}{}
    \footnotetext{$\star$ Authors equally contributed.}
    \footnotetext{Email to:
    \{seokhee.hong, sunkyoung.kim, jinsik.lee\}@lgresearch.ai, \
    guijin.son@onelineai.com
    }
    \let\thefootnote=\oldthefootnote
}

\title{From \redux~to \textsc{Pro}: \\ A Professional Korean Benchmark Suite for LLM Evaluation}

\author{
    Seokhee Hong$^{1, \star}$  \quad
    Sunkyoung Kim$^{1, \star}$ \quad
    \textbf{Guijin Son$^{2}$} \\
    \textbf{Soyeon Kim$^{1}$} \quad
    \textbf{Yeonjung Hong$^{1}$} \quad
    \textbf{Jinsik Lee$^{1}$} \\
    $^1$LG AI Research \quad $^2$OnelineAI
}

\begin{document}

\maketitle
\begin{abstract}
The development of Large Language Models (LLMs) requires robust benchmarks that encompass not only academic domains but also industrial fields to effectively evaluate their applicability in real-world scenarios. In this paper, we introduce two Korean expert-level benchmarks. \textbf{\redux}, reconstructed from the existing \original~\citep{son2024kmmlumeasuringmassivemultitask}, consists of questions from the Korean National Technical Qualification exams, with critical errors removed to enhance reliability. \textbf{\pro} is based on Korean National Professional Licensure exams to reflect professional knowledge in Korea. Our experiments demonstrate that these benchmarks comprehensively represent industrial knowledge in Korea. We release our dataset publicly available.\footnote{\url{https://huggingface.co/datasets/LGAI-EXAONE/KMMLU-Redux}

\url{https://huggingface.co/datasets/LGAI-EXAONE/KMMLU-Pro}}
\end{abstract}

\section{Introduction}

\correspondingfootnote

\input{section/introduction}

\section{\redux}
\label{sec:kmmlu_redux}
\input{section/redux}

\section{\pro}
\label{sec:kmmlu_pro}
\input{section/pro}

\section{Experiments}
\label{sec:experiments}
\input{section/experiments}

\section{Results}
\label{sec:results}
\input{section/results}

\section{Analysis}
\label{sec:analysis}
\input{section/analysis}

\section{Related Works}
\label{sec:related_works}
\input{section/related_works}

\section{Conclusion}
\label{sec:conclusion}
\input{section/conclusion}

\section*{Limitations}
Our benchmarks are limited to text-only and multiple-choice questions for text-only LLMs. It restricts its coverage of real-world licensure exams. Many real-word professional qualification exams include non-textual modalities or require constructed responses such as essay. Our benchmark cannot fully assess all aspects of professional competence or reasoning required in such exams. Expanding to multimodal inputs and open-ended question formats is an important direction for future work.

\section*{Ethical Statements}
All data used in our benchmarks are either publicly available or collected from official licensing materials released by government or professional institutions. For quality control, we hired human annotators to review parsed questions from PDF; they were compensated over the mininum wage in Korea. Our benchmarks would be released under CC-BY-NC-ND 4.0 license.

\bibliography{custom}

\newpage
\appendix

\input{section/appendix}

\end{document}

%% file: section/introduction.tex
As LLMs continue to achieve strong performance across a wide range of subjects~\citep{openai2024openaio1card,gemini2flash,deepseekai2025deepseekr1incentivizingreasoningcapability,research2025exaonedeepreasoningenhanced}, the demand for comprehensive benchmarks has grown.
MMLU~\citep{hendrycks2021measuring} is widely used for its broad coverage of general knowledge from elementary to college level. However, its publicly available online problems has raised concerns about reliability and potential data contamination~\citep{gema2025mmlu, vendrow2025largelanguagemodelbenchmarks,zhao2024mmlucfcontaminationfreemultitasklanguage}.

We identify similar issues in \original~\citep{son2024kmmlumeasuringmassivemultitask}, a widely used benchmark for evaluating Korean expert-level knowledge. The dataset was constructed by crawling websites that provide questions from various exams. We observe noisy samples, including problems that explicitly reveal the answer or non-existent reference, which can mislead performance evaluation. Additionally, we find evidence of contamination between the train and test splits, as well as with common web corpus such as FineWeb2~\citep{penedo2024fineweb-2}.

Instead of collecting data from online sources directly, recent challenging benchmarks~\cite{srivastava2023beyond,rein2024gpqa,phan2025humanitysexam,kazemi2025bigbenchextrahard,pteam2025supergpqascalingllmevaluation} have been constructed through problems and answers authored by human expert.
Although this approach ensures high-quality, contamination-free benchmarks, it is costly to construct and maintain. The high construction cost hinders regular updates~\cite{white2025livebench, jain2025livecodebench}, leaving the benchmarks vulnerable to deprecation and contamination.

Furthermore, existing benchmarks primarily focus on academic knowledge and often overlook the practical applicability of models in industrial or professional contexts. As LLMs are increasingly adopted in industrial domains~\citep{10822885}, it becomes essential to assess whether they possess the necessary expertise to support tasks that require certified knowledge. For example, before deploying an LLM as a legal assistant, one must ensure that the model can reliably understand to meet professional certification standards.

\input{resources/fig_main_figures}

We introduce two benchmarks to address the limitations above and incorporate professional knowledge to evaluate the practical applicability of LLMs.
First, \textbf{\redux}, a refined subset of \original~with 2,587 problems, is built through manual examination by authors to reduce errors and contamination within \original. While the \original~contains wide range of problems, even in high-school level, \redux~only selects Korean National Technical Qualification (KNTQ) exams as sources. The exams require applicants to have either a bachelor's degree or at least nine years experience in industrial field, thus making the benchmark more challenging.

Second, we build \textbf{\pro}, a new challenging benchmark, which consists of 2,822 problems from acquisition exams for Korean National Professional Licensure (KNPL), representing highly specialized professions in Korea.
We include 14 professions (Table~\ref{tab:pro_stats}) from diverse domains.
Unlike \original~that crawls websites, we collect data directly from the official source of each license. After that, human annotators manually examine it to avoid noises. \pro~only includes exams held in the most recent year and would be updated annually with the latest exam to maintain long-term reliability and prevent contamination. See Appendix~\ref{appendix:example_of_redux_and_pro} for examples of each benchmark.

We conduct extensive evaluations of various LLMs on the two benchmarks. Our benchmarks, based on real-world exams, enable aligned analysis with industrial and professional qualifications, effectively revealing the practical strengths of each model. 
As shown in Figure~\ref{fig:main}, \redux{} (left) covers a wide range of industrial domains, allowing it to evaluate the breadth of industrial knowledge. LLMs show robust performance in engineering domains but exhibit notable declines in specialized fields such as Mining \& Resources and Architecture. On the other hand, \pro{} (right) focuses on professional licensure exams and thus assesses whether a model can pass the exams required for high-stakes professions in Korea.
The results show that, in \pro{}, several state-of-the-art models perform strongly in the medicine domain, meeting the passing criteria of most licenses, yet nearly fail in law-related licenses. 

Moreover, we observe the significant performance gaps between our datasets and merely translated datasets like MMMLU~\citep{mmmlu}, especially in domains such as laws, where in-depth knowledge of specific countries is required (see Section~\ref{subsec:law_results}). We argue that these findings underscore the practicality of our benchmarks for assessing the capabilities of models in professional fields within Korea.

To summarize, our contributions are as follows:
\begin{itemize}[itemsep=2pt, parsep=0pt]
\item We improve the previous benchmark, \original, to construct the refined and compact version of the benchmark, \textbf{\redux}, by correcting various errors.
\item We introduce \textbf{\pro}, a new benchmark designed to evaluate high-level professional knowledge in Korea. By imitating real-world license acquisition systems, \pro~assesses the industrial practicality across various professions in Korea.
\item We comprehensively analyze the results of two benchmarks, highlighting the importance of benchmarks specialized in Korea-specific professional knowledge.
\end{itemize}

%% file: resources/fig_main_figures.tex
\begin{figure*}[!tbp]
  \centering
  \resizebox{\textwidth}{!}{
  
  \begin{minipage}[c]{0.45\linewidth} 
    \centering
    \includegraphics[height=1.1\linewidth,width=\linewidth]{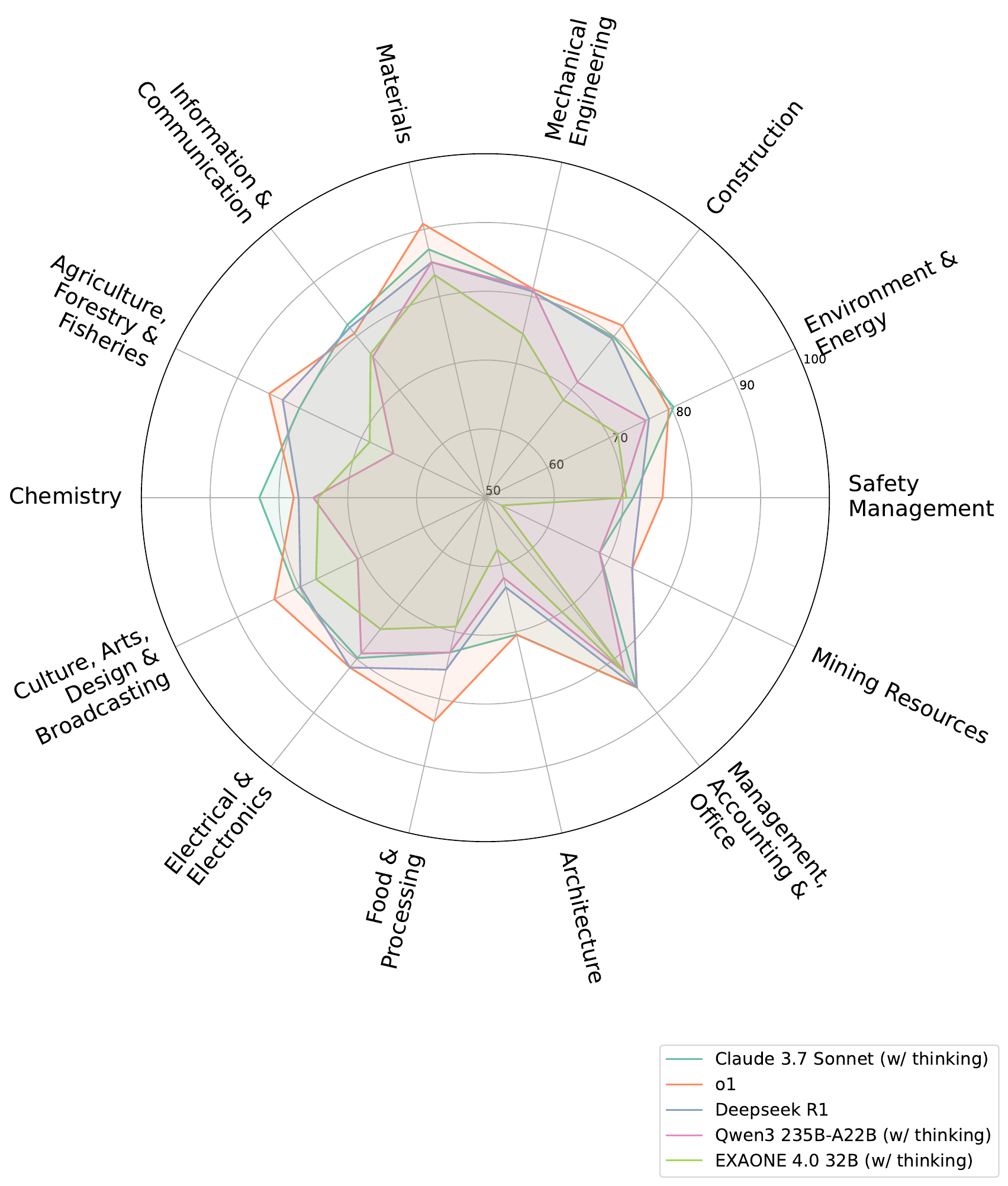}
  \end{minipage}%
  \hspace{\dimexpr0.01\linewidth}
  \begin{minipage}[c]{0.55\linewidth}  
    \centering
    \includegraphics[height=\linewidth,width=\linewidth]{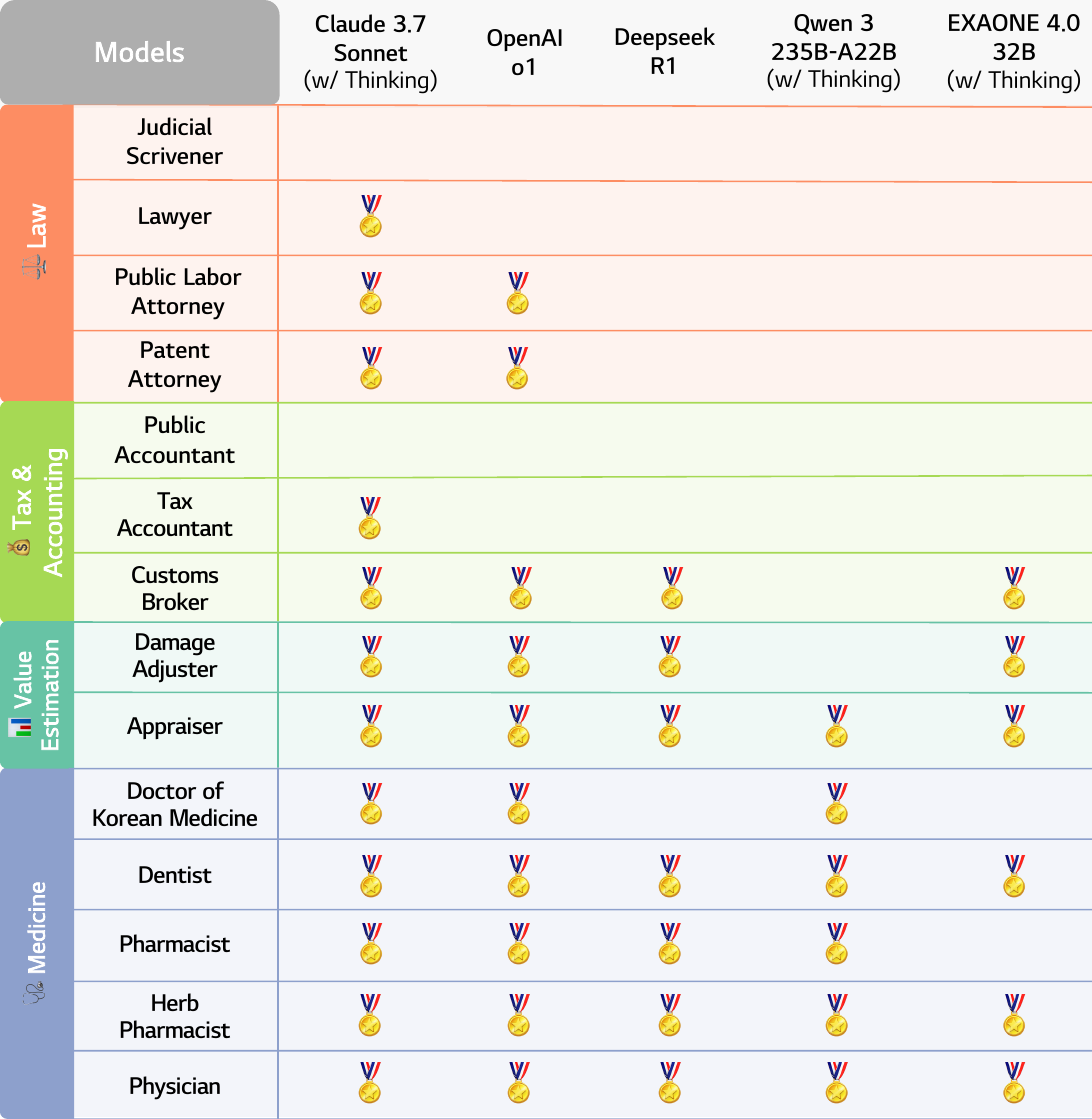}
  \end{minipage}
}
  \caption{Performance of leading reasoning models developed by diverse groups on industrial knowledge for \redux{} (\textbf{left}) and licensure exam pass status (indicated by \includegraphics[height=1em]{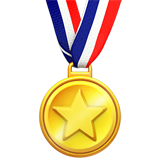}) for \pro{} (\textbf{right}).}
  \label{fig:main}
  \vspace{-1pt}
\end{figure*}

%% file: section/redux.tex
We first revisit \original~\citep{son2024kmmlumeasuringmassivemultitask} to examine its quality. Building on these insights, we construct \redux, a cleaned and compact version of the \original. We carefully denoise against the \original~and increase difficulty.

\subsection{Revisiting \original}\label{subsec:revisiting}
\original~plays a significant role in the NLP community as a de facto standard for evaluating LLMs on Korea-specific expert knowledge~\citep{yoo2024hyperclovaxtechnicalreport,research2024exaone35serieslarge,kananallmteam2025kananacomputeefficientbilinguallanguage}. 
The dataset was constructed by crawling websites\footnote{https://www.kinz.kr/} where various exam questions are uploaded by people online, spanning from high school to professional qualification exams.
Upon closer inspection, we identify several noises and limitations, which can be categorized into three types: 1) duplication issues, 2) dataset errors, and 3) contamination.

\paragraph{Duplication Issue}
As the \original~crawls problems from hundreds of exams in Korea, we observe multiple duplicated questions across related exams.
Notably, duplicated samples occur not only within the test set but also between the training and test sets.
Using the Longest Common Sequence (LCS) algorithm to investigate overlaps, we could find 5.36\% duplication within the test set and a 5.46\% contamination between train and test set.

\input{resources/fig_error_types_acc}

\paragraph{Dataset Errors}
Following \citet{gema2025mmlu}, we investigate the extent to which various error types appear in the \original~test set and their potential impact on LLM performance. We identify four representative error types: leaked answers, ill-posed questions, poor clarity, and notation errors. We then annotate the entire test set using GPT-4o\footnote{gpt-4o-2024-11-20}~\citep{openai2024gpt4ocard}. In total, we find that 7.66\% of the data contains one of the errors mentioned above. We describe the details of the investigation of dataset errors in Appendix~\ref{appendix:detail_redux_errors}.

In Figure~\ref{fig:error_type_acc}, we observe significant discrepancies in the LLM's performance between erroneous and correct samples.
Notably, the performance of instances with leaked answers drops significantly when the LLMs are assessed on the clean dataset. Conversely, all models’ scores increase on ill-posed questions, underscoring their inability to identify the correct answer for poorly formulated questions.

\paragraph{Contamination}
The \original~was primarily sourced online, making them highly susceptible to contamination from web-crawled training corpora.
When applying n-gram contamination detection~\citep{lambert2025tulu3pushingfrontiers,grattafiori2024llama3herdmodels} to the Korean subset of FineWeb2~\citep{penedo2024fineweb-2} and the \original~dataset, 1.88\% of the data were flagged as contaminated.

\subsection{Dataset Construction}\label{subsec:dataset_construction_redux}

\original~consists of approximately $35k$ examples, making evaluation resource-intensive. To reduce this burden, we first restrict the scope to a subset of high-difficulty exams (Section~\ref{subsubsec:challenging_exams}). Furthermore, to ensure the reliability of the benchmark, we manually conduct a thorough examination of the dataset to eliminate errors (Section~\ref{subsubsec:denosing}).

\input{resources/tab_kmmlu_pro_stats}

\subsubsection{Filtering Non-Challenging Problems}\label{subsubsec:challenging_exams}
Since \original~includes a variety of exams in Korea, spanning from high school to professional certification exams, we filter out the easier exams to build a more challenging benchmark.
Specifically, we choose Korean National Technical Qualification (KNTQ) exams, which are primarily designed to assess practical technical competencies required in industrial field.

The qualifications require applicants to have either a bachelor's degree or at least nine years of professional experience. We adopt a collection of 100 KNTQ exams across the 14 domains in total. We only include the most recent exam for each qualification, thereby avoiding outdated knowledge being evaluated\footnote{We have compiled a list of all KNTQs exams alongside their most recent exam dates in Appendix~\ref{appendix:redux_KNTQ_list}. Our aim is to make these available to LLM researchers and developers to help prevent data contamination.}.

To further discriminate simple problems, we leverage the performances of LLMs on the data~\citep{wang2024mmlupro,zellers-etal-2019-hellaswag,lee-etal-2023-square}. By employing seven smaller LLMs\footnote{Llama 3.2 3B~\citep{llama3.2}, Qwen 2.5 3B~\citep{qwen2025qwen25technicalreport}, Gemma 3 4B IT~\citep{gemma3}, Kanana Nano 2.1B Instruct~\citep{kananallmteam2025kananacomputeefficientbilinguallanguage}, EXAONE 3.5 2.4B~\citep{research2024exaone35serieslarge}, DeepSeek-R1-Distill-Qwen-1.5B~\citep{deepseekai2025deepseekr1incentivizingreasoningcapability} EXAONE Deep 2.4B~\citep{research2025exaonedeepreasoningenhanced}, and Ko-R1-7B-v2.1~\citep{ko-r1-7b-v2.1}.}, we mark a data as \textit{easy} if four or more models correctly predict its answer. Through this process, we remove 38.6\% of the dataset.
 
\subsubsection{Denoising}\label{subsubsec:denosing}
To remove noises, we follow the processes described in Section~\ref{subsec:revisiting}. Specifically,  we first manually review the dataset to minimize errors. Next, we perform decontamination to prevent potential data leakage from pre-training corpora. Additionally, we detect inner duplication and against the training and test sets in \original, thus finally remove all duplicates. 

\subsubsection{Final Statistics}
For \redux, we have collected 2,587 problems from 100 KNTQ exams. Among these, 596 problems are from exams that require over nine years of professional experience to acquire the qualification. To categorize the dataset into 14 domains, we follow the Korean Standard Industrial Classification (KSIC) published by Statistics Korea\footnote{https://www.kostat.go.kr/} as the qualification system is primarily designed to align with industrial fields. Figure~\ref{fig:redux_category} in Appendix~\ref{appendix:redux_KNTQ_list} illustrates the distribution of domain of \redux.

%% file: resources/fig_error_types_acc.tex
\begin{figure}
    \centering
    \includegraphics[width=\linewidth]{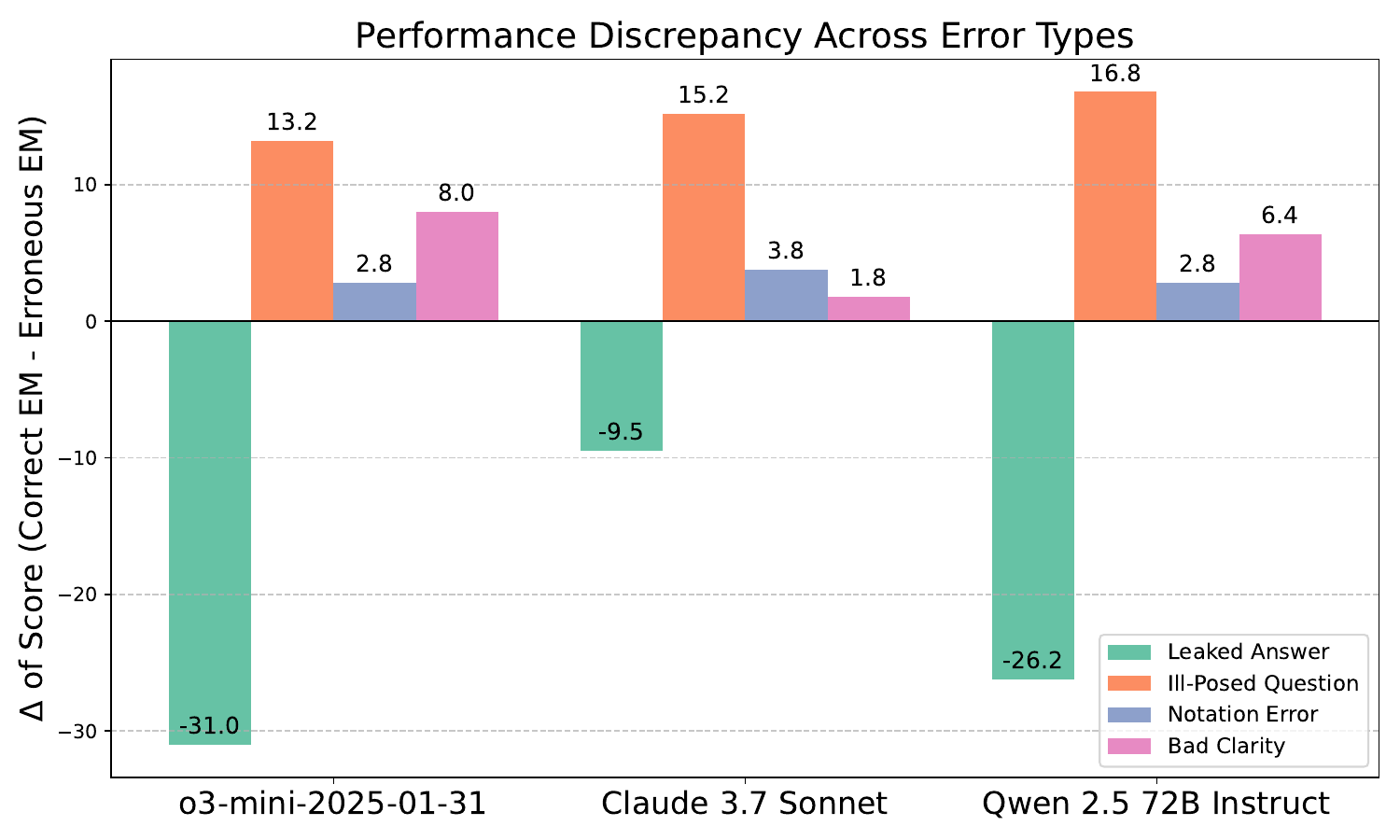}
    \caption{Performance differences in LLMs on the erroneous versus correct dataset. Leaked answer errors tend to overestimate model capabilities, while three other error types hinder LLMs to correctly predict true answers.}
    \label{fig:error_type_acc}
\end{figure}

%% file: resources/tab_kmmlu_pro_stats.tex
\begin{table*}[ht]
\centering
\resizebox{\textwidth}{!}{
\begin{tabular}{@{}lllc@{}}
\toprule
Domain & Names of KNPLs & U.S. Equivalent & \# of Instances \\ \midrule
\multirow{4}{*}{Law}  & Certified Judicial Scrivener & Paralegal, Legal Document Assistant, or Notary Public (no direct equivalent) & 198 \\
 & Lawyer~\citep{kimyeeun-etal-2024-developing} & Attorney-at-Law & 150 \\
& Certified Public Labor Attorney & Labor \& Employment Lawyer (requires J.D. and bar admission; no separate certification in the U.S.) & 239 \\
 & Certified Patent Attorney & Patent Attorney (JD + USPTO registration required) & 109 \\ \midrule
\multirow{3}{*}{Tax \& Accounting} & Certified Public Accountant (CPA) & Certified Public Accountant (CPA) – Exact Equivalent & 208 \\
 & Certified Tax Accountant & Enrolled Agent (IRS) or CPA with Tax Specialization & 238 \\ 
& Certified Customs Broker & U.S. Customs Broker (licensed by U.S. Customs and Border Protection - CBP) & 159 \\ \midrule
\multirow{2}{*}{Value Estimation}  & Certified Damage Adjuster (CDA) & Claims Adjuster / Insurance Adjuster (state-licensed) & 120 \\
 & Certified Appraiser & Certified Real Estate Appraiser (licensed at the state level) & 196 \\ \midrule
 \multirow{5}{*}{Medicine}
 & Doctor of Korean Medicine & Licensed Acupuncturist (L.Ac.) or Doctor of Acupuncture and Oriental Medicine (D.A.O.M.) & 288 \\
 & Dentist~\citep{kweon2024kormedmcqamultichoicequestionanswering} & Doctor of Dental Surgery (D.D.S.) / Doctor of Dental Medicine (D.M.D.) & 252 \\
 & Pharmacist~\citep{kweon2024kormedmcqamultichoicequestionanswering} & Doctor of Pharmacy (Pharm.D.) & 271 \\
 & Herb Pharmacist & Herbalist (non-licensed or CAM-certified depending on state) & 244 \\ 
  & Physician~\citep{kweon2024kormedmcqamultichoicequestionanswering} & Medical Doctor (M.D./D.O.) & 150 \\ \midrule
 & Total &  & \textbf{2822} \\ \bottomrule
\end{tabular}%
}
\caption{The list of National Professional Licenses (NPLs) used for \pro~and their corresponding statistics. The names of NPLs are translated from those in Korea and we also report equivalent licences in U.S. We use KorMedMCQA~\citep{kweon2024kormedmcqamultichoicequestionanswering} for three licenses in the Medical category, and KBL~\citep{kimyeeun-etal-2024-developing} for the bar exam of lawyer.}
\label{tab:pro_stats}
\end{table*}

%% file: section/pro.tex
A major challenge in building benchmarks from online sources is data contamination~\citep{zhao2024mmlucfcontaminationfreemultitasklanguage,jain2025livecodebench,roberts2024to}. While some studies address this by having experts manually create problems~\cite{srivastava2023beyond,rein2024gpqa,phan2025humanitysexam,kazemi2025bigbenchextrahard, pteam2025supergpqascalingllmevaluation}, this approach is costly and time-consuming. As an alternative, recent work explores periodic releases of fresh subsets.~\citep{white2025livebench,jain2025livecodebench}. 

Motivated by these approaches, we focus on the Korean National Professional Licensure (KNPL) exams. These are high-stakes exams administered annually that pose a significant challenge. Unlike benchmarks crafted by a small set of experts~\cite{rein2024gpqa,phan2025humanitysexam}, our approach leverages the well-established curricula of professional licensing systems, designed to assess real-world professional knowledge.

\subsection{Korean National Professional Licensures}
We choose KNPL exams for main source of \pro. KNPL exams target high-level professionals, such as lawyers, accountants, or physicians, requiring advanced knowledge, critical reasoning, and ethical judgment.
Among them, we select 14 KNPLs representing highly specialized and regulated professions in Korea
(See Table~\ref{tab:pro_stats} for the list of KNPLs and their equivalent licensure in U.S.). These licenses are legally mandated credentials required to practice in their respective domains. As such, they serve as institutionalized gateways to high-status occupations with significant entry barriers.
Our evaluation simulates real-world assessment standards by incorporating official exam pass criteria, aligning model performance with human standards.

\subsection{Dataset Collection and Annotation}
In Korea, the government releases and manages the questions for KNPL acquisition exams. We directly download the PDF files from the government's websites for each license and use GPT-4o~\citep{openai2024gpt4ocard} for OCR parsing. As our dataset sources from the official PDFs, we can enhance the quality of the dataset, avoiding potential errors when collecting from online text~\citep{pteam2025supergpqascalingllmevaluation}. Since GPT-4o has difficulty handling tables and low-resolution PDFs, we employ human annotators to review parsed questions. When a problem contains an image, the annotators convert it into text that conveys the same meaning of the image, if possible.
Notably, our process remains relatively cost-efficient as it requires human annotators solely for error reviewing tasks, not full annotation. We demonstrate detailed annotation process in Appendix~\ref{appendix:annotation_guideline}

We follow previous works, releasing a new set of questions periodically~\cite{white2025livebench, jain2025livecodebench}. Since the exams are conducted annually, we commit to collect and release questions from the exams held just before the current year.

\subsection{Decontamination}
We also adopt the same process outlined in Section~\ref{subsubsec:denosing} to ensure \pro~is free from contamination. When conducting n-gram match between \pro, FineWeb2, and the training and validation sets of \original, we did not find any contaminated examples. As a result, we can retain all 2,822 data points in the \pro, maintaining its contamination-free integrity.

%% file: section/experiments.tex
\input{resources/tab_main_results}

We select a diverse set of baseline models varying in size, multilingual capability, and reasoning ability. By default, we apply a zero-shot Chain-of-Thought (CoT)~\citep{NEURIPS2022_8bb0d291} prompt written in Korean. However, in early experiments, we observed that some reasoning models perform worse when prompted in Korean. Therefore, we evaluate those models using both Korean and English prompts and report the scores with the highest average score across \redux{} and \pro{}\footnote{For further details, please see Section~\ref{subsec:prompt_language}.}.
We use greedy decoding for non-reasoning models, while for reasoning models, we adopt the temperature of 0.6 and top-p~\citep{Holtzman2020The} of 0.95. For more details, see Appendix~\ref{appendix:evaluation_details}.

Evaluating open-weight models was conducted over four days using sixteen NVIDIA A100 GPUs. For closed models, we accessed them via API calls, which incurred a total cost of approximately \$4,000.

\paragraph{Metrics} In addition to accuracy, our primary evaluation metric, we also report the number of licensure exams each LLM passes in \pro. To better align with human evaluation standards, our procedure is designed to mirror the official certification standards. However, for licenses that rely heavily on image‐based questions, full replication is not possible. See Appendix~\ref{appendix:pro_scoring} for details.

%% file: resources/tab_main_results.tex
\begin{table*}[!t]
\centering
\footnotesize
\resizebox{\textwidth}{!}{
\begin{tabular}{@{}lcccc@{}}
\toprule
\multicolumn{1}{l|}{} & KMMLU-Redux & \multicolumn{2}{c|}{KMMLU-Pro} &  \\
\multicolumn{1}{l|}{} & Acc & Acc & \multicolumn{1}{c|}{\# of passed KNPLs} & Avg. Acc (micro) \\ \midrule
\multicolumn{5}{l}{Open-weight Models} \\ \midrule
\multicolumn{1}{r|}{ Aya Expanse 32B~\citep{dang2024ayaexpansecombiningresearch}} & 33.05 & 31.26 & \multicolumn{1}{c|}{0/14} & 32.12 \\
\multicolumn{1}{r|}{ Gemma 3 12B IT~\citep{gemma3}} & 46.70 & 45.82 & \multicolumn{1}{c|}{2/14} & 46.24 \\
\multicolumn{1}{r|}{ Phi-4 (14B)~\citep{abdin2024phi4technicalreport}} & 49.75 & 45.32 & \multicolumn{1}{c|}{1/14} & 47.44 \\
\multicolumn{1}{r|}{ Mistral Small 3.1 Instruct (24B)~\citep{mistralsmall3.1}} & 52.92 & 49.49 & \multicolumn{1}{c|}{3/14}  & 51.13 \\
\multicolumn{1}{r|}{ Gemma 3 27B IT~\citep{gemma3}} & 54.04 & 51.03 & \multicolumn{1}{c|}{2/14}  & 52.47 \\
\multicolumn{1}{r|}{  Llama 3.3 70B Instruct~\citep{llama3.3}} & 56.17 & 53.24 & \multicolumn{1}{c|}{3/14} & 54.64 \\
\multicolumn{1}{r|}{ Qwen3-14B~\citep{yang2025qwen3technicalreport}} & 57.25 & 53.02 & \multicolumn{1}{c|}{3/14} & 55.04 \\
\multicolumn{1}{r|}{ Qwen3-30B-A3B~\citep{yang2025qwen3technicalreport}} & 58.41 & 52.33 & \multicolumn{1}{c|}{3/14} & 55.24 \\
\multicolumn{1}{r|}{ C4AI Command A (111B)~\citep{c4aicommanda}} & 62.93 & 57.48 & \multicolumn{1}{c|}{3/14} & 60.07 \\
\multicolumn{1}{r|}{ Qwen3-32B~\citep{yang2025qwen3technicalreport}} & 64.98 & 58.86 & \multicolumn{1}{c|}{3/14} & 61.79 \\
\multicolumn{1}{r|}{ EXAONE 4.0 32B~\citep{research2025exaone40unifiedlarge}} & 64.79 & 60.01 &  \multicolumn{1}{c|}{3/14} & 62.30 \\
\multicolumn{1}{r|}{ Llama-4-Scout-17B-16E-Instruct~\citep{llama4}} & 67.49 & 58.14 & \multicolumn{1}{c|}{4/14} & 62.61 \\
\rowcolor[rgb]{0.9,0.9,0.9} \multicolumn{1}{r|}{ Qwen3-30B-A3B~{\smaller[2]~(w/ thinking)}~\citep{yang2025qwen3technicalreport}} & 65.25 & 60.52 & \multicolumn{1}{c|}{3/14} & 62.78 \\
\rowcolor[rgb]{0.9,0.9,0.9} \multicolumn{1}{r|}{ Qwen3-14B~{\smaller[2]~(w/ thinking)}~\citep{yang2025qwen3technicalreport}} & 65.71 & 60.18 & \multicolumn{1}{c|}{2/14} & 62.82 \\
\multicolumn{1}{r|}{ DeepSeek V3 (671B)~\citep{deepseekai2025deepseekv3technicalreport}} & 65.64 & 60.77 & \multicolumn{1}{c|}{4/14} & 63.10 \\
\rowcolor[rgb]{0.9,0.9,0.9} \multicolumn{1}{r|}{ Qwen3-32B~{\smaller[2]~(w/ thinking)}~\citep{yang2025qwen3technicalreport}} & 68.77 & 61.14 & \multicolumn{1}{c|}{3/14} & 64.79 \\
\rowcolor[rgb]{0.9,0.9,0.9} \multicolumn{1}{r|}{ QwQ 32B~\citep{qwq32b}} & 67.34 & 63.94 & \multicolumn{1}{c|}{5/14} & 65.57 \\
\multicolumn{1}{r|}{ Qwen3-235B-A22B~\citep{yang2025qwen3technicalreport}} & 69.54 & 62.12 & \multicolumn{1}{c|}{4/14} & 65.67 \\
\rowcolor[rgb]{0.9,0.9,0.9} \multicolumn{1}{r|}{ EXAONE 4.0 32B~{\smaller[2]~(w/ thinking)}~\citep{research2025exaone40unifiedlarge}} & 72.71 & 67.67 &  \multicolumn{1}{c|}{6/14} & 70.08 \\
\rowcolor[rgb]{0.9,0.9,0.9} \multicolumn{1}{r|}{ Qwen3-235B-A22B{\smaller[2]~(w/ thinking)}~\citep{yang2025qwen3technicalreport}} & 74.49 & 68.22 & \multicolumn{1}{c|}{6/14} & 71.22 \\
\multicolumn{1}{r|}{ Llama-4-Maverick-17B-128E-Instruct~\citep{llama4}} & 77.58 & 68.10 & \multicolumn{1}{c|}{4/14} & 72.63 \\
\rowcolor[rgb]{0.9,0.9,0.9} \multicolumn{1}{r|}{ DeepSeek R1 (671B)~\citep{deepseekai2025deepseekr1incentivizingreasoningcapability}} & 78.51 & 71.33 & \multicolumn{1}{c|}{7/14} & 74.76 \\
\midrule

\multicolumn{5}{l}{Closed Models} \\
\midrule
\multicolumn{1}{r|}{ GPT-4.1 mini~{\smaller[2]~(2025-04-14)}~\citep{gpt4.1}} & 67.03 & 62.18 & \multicolumn{1}{c|}{4/14} & 64.50 \\
\rowcolor[rgb]{0.9,0.9,0.9} \multicolumn{1}{r|}{ o3-mini~{\smaller[2]~(2025-01-31)}~\citep{o3mini}} & 67.84 & 62.05 & \multicolumn{1}{c|}{3/14} & 64.82 \\
\rowcolor[rgb]{0.9,0.9,0.9} \multicolumn{1}{r|}{ Grok-3-mini-beta~\citep{grok3}} & 71.47 & 65.08 & \multicolumn{1}{c|}{5/14} & 68.14 \\
\multicolumn{1}{r|}{ Grok-3-beta~\citep{grok3}} & 72.90 & 68.37 & \multicolumn{1}{c|}{7/14} & 70.54 \\
\rowcolor[rgb]{0.9,0.9,0.9} \multicolumn{1}{r|}{ o4-mini~{\smaller[2]~(2025-04-16)}~\citep{o3}} & 75.80 & 69.65 & \multicolumn{1}{c|}{6/14} & 72.59 \\
\multicolumn{1}{r|}{ GPT-4.1~{\smaller[2]~(2025-04-14)}~\citep{gpt4.1}} & 75.86 & 72.99 & \multicolumn{1}{c|}{10/14} & 74.36 \\
\multicolumn{1}{r|}{ Claude 3.7 Sonnet~\citep{claude3.7}} & 76.88 & 74.52 & \multicolumn{1}{c|}{10/14} & 75.65 \\
\rowcolor[rgb]{0.9,0.9,0.9} \multicolumn{1}{r|}{ o3~\citep{o3}} & 79.92 & 73.60 & \multicolumn{1}{c|}{9/14} & 76.62 \\
\rowcolor[rgb]{0.9,0.9,0.9} \multicolumn{1}{r|}{ Claude 3.7 Sonnet~{\smaller[2]~(w/ thinking)}~\citep{claude3.7}} & 79.36 & 77.70 & \multicolumn{1}{c|}{12/14} & 78.49 \\
\rowcolor[rgb]{0.9,0.9,0.9} \multicolumn{1}{r|}{ o1~\citep{openai2024openaio1card}} & 81.14 & 78.09 & \multicolumn{1}{c|}{10/14} & 79.55 \\ \bottomrule

\end{tabular}%
}
\caption{The main evaluation results of \redux~and \pro~benchmarks on various LLMs. \colorbox{gray!20}{The gray-shaded models} stand for reasoning models. 
The results of models with size \textless~10B are presented in Table~\ref{tab:main_results_smaller} in Appendix~\ref{appendix:main_result_smaller}.
}
\label{tab:main_results}
\end{table*}

%% file: section/results.tex
\subsection{Main Results}
Table~\ref{tab:main_results} presents the overall performance on \redux~and \pro. The o1 model achieves the highest average accuracy (79.55), followed by Claude 3.7 with Thinking (78.49). Among open-weight models, DeepSeek's R1 even outperforms many closed models. Models equipped with reasoning capabilities consistently perform better than their non-reasoning models, such as the Qwen3 series and EXAONE 4.0.

Beyond accuracy, we evaluate each model's ability to pass the licensure exams under the official criteria for \pro. Specifically, in most licenses, a model must score at least 40\% in each subject and achieve an overall average of 60\% to pass. Claude 3.7 with Thinking succeeds in 12 out of 14 KNPL licensure exams, the highest among all evaluated models. In contrast, although the o1 model achieves higher accuracy, it qualifies for fewer licenses than Claude 3.7 with Thinking. This highlights the importance of balanced competence across subjects, as required in real-world certification exams.

\subsection{Performance Across Industrial Domains in \redux}
Through \redux, we assess the technical competencies of models across diverse industrial fields. 
As shown in Figure~\ref{fig:main} (left), models demonstrate varying levels of performance across these fields. For example, 
leading models with reasoning capabilities struggle in domains such as Safety Management, Mining Resources, and Architecture, achieving under 80\% accuracy, which highlights persistent challenges in underrepresented or highly specialized fields. 
In contrast, the models obtain scores exceeding 80\% in Materials and Management, Accounting \& Office areas.
The full results for all models across 14 domains are presented in Appendix~\ref{appendix:breakdown_redux}.

\input{resources/fig_law_results}

\subsection{Professional Licensure Exam Performance on \pro}
We provide a breakdown of the \pro~results by analyzing which licensure exams are most frequently qualified by LLMs. Figure~\ref{fig:main} (right) shows the pass rates of LLMs across 14 KNPLs. While many models pass the exams in the medicine domain, most fail in Law and Tax \& Accounting, with only DeepSeek R1 and EXAONE 4.0 32B passing the Customs Broker exam among open-weight models. Notably, no models pass the Judicial Scrivener or Public Accountant exams. This trend becomes even more evident in the full results across a broader range of LLMs; the only licenses that relatively smaller models (\textless 20B) are able to pass in the medicine domain (see Table~\ref{tab:pro_category_result_full} in Appendix~\ref{appendix:breakdown_pro}). 

Moreover, many LLMs fail in the licensure exams even when scoring above 60\%; for example, o3-mini, Qwen3-235B-A22B, and Llama-4-Maverick score over 85\% on the Pharmacist exam but still fail to qualify due to not meeting the threshold of law-related subject in the exam. These cases highlight the difficulty of acquiring region-specific domain knowledge, particularly in legal subjects governed by Korean law.

%% file: resources/fig_law_results.tex
\begin{figure*}[ht!]
  \centering
  \includegraphics[width=\textwidth]{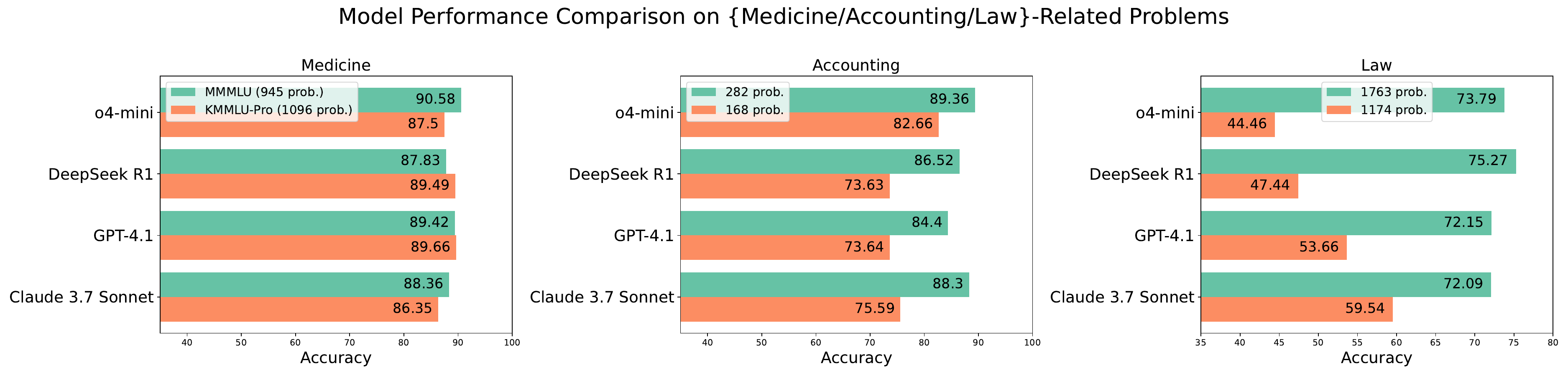}
    \caption{Performance of four LLMs on \{Medical\textbf{(left)}, Accounting\textbf{(center)}, Law\textbf{(right)}\}-relevant subsets from the MMMLU (Korean)~\citep{mmmlu} and \pro. While the discrepancies in scores are narrow in the medicine domain, they are wider in law-related problems, emphasizing the need for datasets that reflecting real professional knowledge in Korea.
    }
    \label{fig:law_results}
\end{figure*}

%% file: section/analysis.tex
\subsection{The Importance of Locally Adapted Benchmarks}\label{subsec:law_results}

To highlight the importance of evaluation grounded in local context~\citep{plaza2024spanishllmbenchmarksmmlu,singh2025globalmmluunderstandingaddressing}, we compare category-level performance between benchmarks translated from English and \pro. Specifically, we focus on subjects related to law, accounting, and medicine,~\footnote{\{\textit{professional\_law}, \textit{jurisprudence}, \textit{international\_law}\} for Law. \{\textit{professional\_accounting}\} for Accounting. \{\textit{professional\_medicine}, \textit{clinical\_knowledge}, \textit{college\_medicine}, \textit{medical\_genetics}, \textit{anatomy}\} for Medicine.} selecting relevant subjects from the Korean subset of MMMLU~\citep{mmmlu}, as well as \pro.

As shown in Figure~\ref{fig:law_results}, the performance gap is relatively small in categories such as medicine, where domain knowledge is largely consistent across countries and cultures. In contrast, categories such as law, where substantial differences in content are expected, show a significantly larger gap. This suggests that MMMLU, which relies on direct translation of law questions based on U.S. standards, cannot adequately represent knowledge of Korean law. These findings highlight the importance of our dataset, which reflects authentic professional knowledge specific to the Korean context.

\input{resources/fig_kmmlu_vs_kmmlu_redux}

\subsection{\original~vs \redux}
Figure~\ref{fig:kmmlu_vs_kmmlu_redux} illustrates the performance of LLMs on \original~and \redux. The LLMs' performances on \redux~is lower than on \original, due to our filtration process which aims to retain only challenging problems from \original~(see Section~\ref{subsubsec:challenging_exams}). Despite this decrease, there is a near-perfect monotonic association between the results, with a Spearman's rank correlation coefficient ($\rho$) of 0.995, suggesting they are highly correlated.

\subsection{Impact of Reasoning Budget} 
\input{resources/fig_thinking_budget}
Recent studies have shown that increasing reasoning efforts can enhance model performance~\citep{muennighoff2025s1simpletesttimescaling,claude3.7,yang2025qwen3technicalreport}. To examine how reasoning \textit{budget} affects performance on each licenses in \pro, we conduct an experiment varying the number of tokens allocated to the reasoning path. We adopt Qwen3-32B~\citep{yang2025qwen3technicalreport} and Claude 3.7 Sonnet~\citep{claude3.7}\footnote{We select the models because they either natively support the “thinking budget” or have reported experimental results on thinking budgets in their technical report.}. For each budget $b \in B$, we generate $n$ different responses per question and average the scores. We set $n=4$ for Qwen3, but $n=2$ for Claude due to the generation cost.

As shown in Figure~\ref{fig:thinking_budget}, we observe a positive correlation between the reasoning budget and the overall score of \pro{} for both models. However, this trend does not hold uniformly across all licenses. 
For example, both models show trivial gains on the Judicial Scrivener and Herb Pharmacist licenses, indicating that more reasoning does not always boost performance.

We further analyze the effect of enabling reasoning itself. To be specific, we compare the scores of reasoning models with their corresponding non-reasoning models on performance in \pro{} licenses, observing how enabling reasoning impacts results. As shown in Table~\ref{tab:reasoning_vs_non_reasoning} in Appendix~\ref{appendix:reasoning_vs_non_reasoning}, reasoning models exhibit different trends across licenses. For example, in the Judicial Scrivener exam, most reasoning models do not show significant performance gains over their non-reasoning counterparts, except for Qwen3-1.7B, which aligns with the results of Figure~\ref{fig:thinking_budget}. In contrast, enabling reasoning boosts performance for many models on CPA exams, demonstrating the impact of reasoning on calculation-intensive tasks.

\subsection{Impact of Prompt Language}\label{subsec:prompt_language}
\input{resources/tab_ko_prompt}
Prompt language can significantly influence model behavior, raising concerns about consistency in multilingual settings~\citep{wang2025polymathevaluatingmathematicalreasoning,zhang2025pmmevalparallelmultilingualmultitask,lai-nissim-2024-mcot}. Since all questions in our datasets are written in Korean, using Korean prompts is a natural choice. However, we observe that some models perform worse when prompted in Korean. Table~\ref{tab:ko_prompt} presents the performance difference between English and Korean prompts. The Llama-4 model series exhibits the most substantial drop in performance\footnote{The Llama-4 models did not follow the prompt but ended their response with "\textit{The best answer is}", even with the Korean prompt.}, while closed models such as Grok-3-mini-beta and o4-mini show minimal change.

%% file: resources/fig_kmmlu_vs_kmmlu_redux.tex
\begin{figure}[th]
    \centering
    \includegraphics[width=\linewidth]{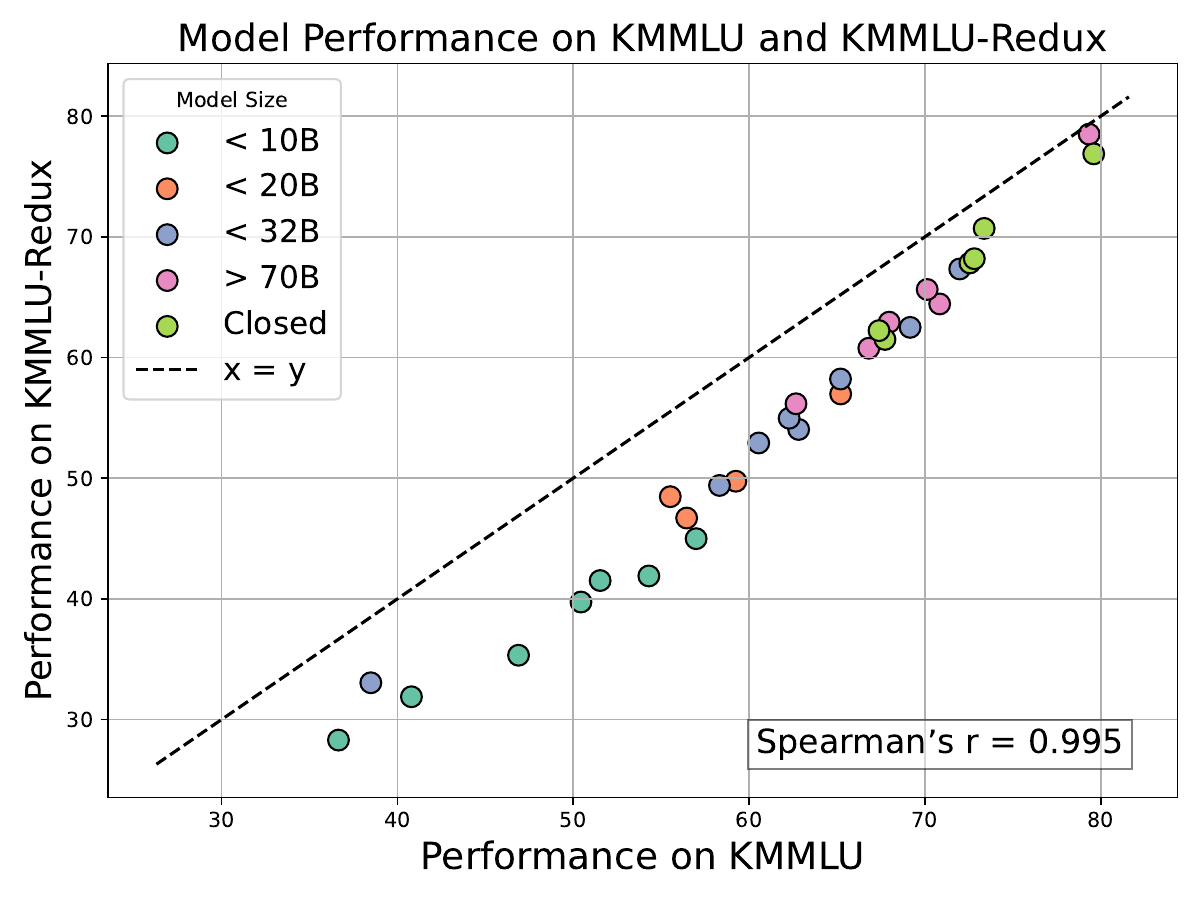}
    \caption{Performance of LLMs on \original~and \redux. A high $\rho$ value indicates a strong correlation between the results of the two benchmarks.
    }
    \label{fig:kmmlu_vs_kmmlu_redux}
\end{figure}

%% file: resources/fig_thinking_budget.tex
\begin{figure*}[ht]
  \centering
  \includegraphics[width=\textwidth]{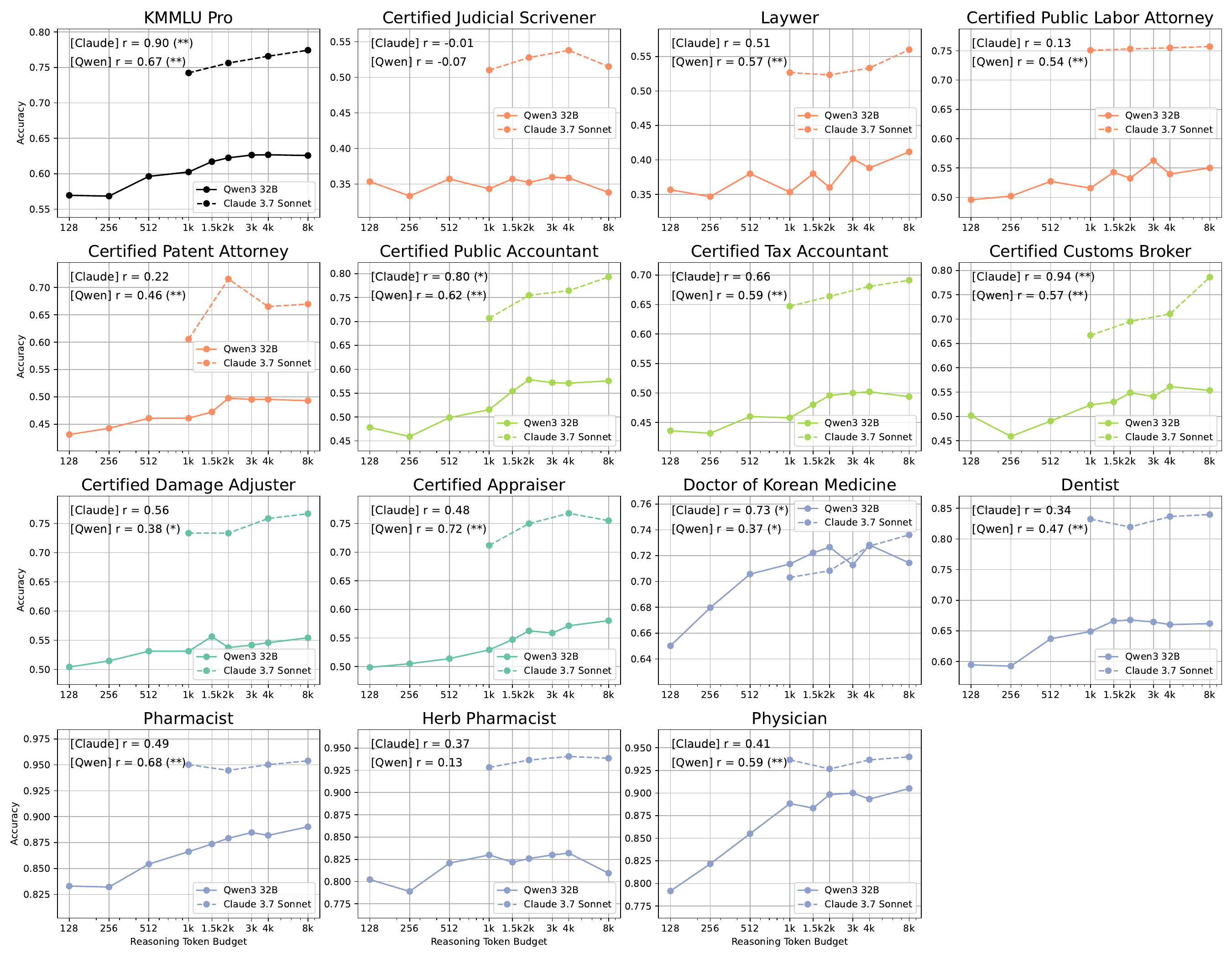}
    \caption{Reasoning budget results of Qwen3-32B and Claude 3.7 Sonnet on \pro. $r$ indicates the Pearson Correlation coefficient value between the reasoning token budget and the accuracy. The responses are sampled multiple times for each thinking budget setting; $n=4$ and $n=2$, respectively, for Qwen and Claude. * and ** denote statistical significance, indicating p-value <0.05 and <0.01, respectively.
    }
    \label{fig:thinking_budget}
\end{figure*}

%% file: resources/tab_ko_prompt.tex
\begin{table}[!ht]
\resizebox{\columnwidth}{!}{%
\begin{tabular}{@{}lcccccc@{}}
\toprule
\multicolumn{1}{c}{} & \multicolumn{3}{c}{\redux} & \multicolumn{3}{c}{\pro} \\
\multicolumn{1}{c}{} & English & Korean & \textit{diff} (\%) & English & Korean & \textit{diff} (\%) \\ \midrule
\multicolumn{1}{l|}{Qwen3-32B~{\smaller[2]~(w/ thinking)}} & 68.77 & 69.08 & \multicolumn{1}{l|}{{\color[HTML]{3166FF} \textbf{+0.5\%}}} & 61.14 & 60.66 & {\color[HTML]{FD6864} \textbf{-0.8\%}} \\
\multicolumn{1}{l|}{o4-mini~{\smaller[2]~(2025-04-16)}} & 75.80 & 76.17 & \multicolumn{1}{l|}{{\color[HTML]{3166FF} \textbf{+0.5\%}}} & 69.65 & 69.10 & {\color[HTML]{FD6864} \textbf{-0.8\%}} \\
\multicolumn{1}{l|}{Qwen3-235B-A22B~{\smaller[2]~(w/ thinking)}} & 74.49 & 75.11 & \multicolumn{1}{l|}{{\color[HTML]{3166FF} \textbf{+0.8\%}}} & 68.22 & 66.98 & {\color[HTML]{FD6864} \textbf{-1.8\%}} \\
\multicolumn{1}{l|}{Grok-3-mini-beta} & 71.47 & 70.85 & \multicolumn{1}{l|}{{\color[HTML]{FD6864} \textbf{-0.9\%}}} & 65.08 & 64.89 & {\color[HTML]{FD6864} \textbf{-0.3\%}} \\
\multicolumn{1}{l|}{Qwen3-14B~{\smaller[2]~(w/ thinking)}} & 65.71 & 65.40 & \multicolumn{1}{l|}{{\color[HTML]{FD6864} \textbf{-0.5\%}}} & 60.18 & 59.48 & {\color[HTML]{FD6864} \textbf{-1.2\%}} \\
\multicolumn{1}{l|}{EXAONE Deep 32B} & 58.33 & 56.17 & \multicolumn{1}{l|}{{\color[HTML]{FD6864} \textbf{-3.7\%}}} & 52.33 & 52.19 & {\color[HTML]{FD6864} \textbf{-0.3\%}} \\ 
\multicolumn{1}{l|}{DeepSeek R1 (671B)} & 78.51 & 75.38 & \multicolumn{1}{l|}{{\color[HTML]{FD6864} \textbf{-4.0\%}}} & 71.33 & 70.62 & {\color[HTML]{FD6864} \textbf{-1.0\%}} \\ 
\multicolumn{1}{l|}{QwQ 32B} & 67.34 & 62.66 & \multicolumn{1}{l|}{{\color[HTML]{FD6864} \textbf{-6.9\%}}} & 63.94 & 59.95 & {\color[HTML]{FD6864} \textbf{-6.2\%}} \\
\multicolumn{1}{l|}{EXAONE Deep 7.8B} & 44.99 & 40.82 & \multicolumn{1}{l|}{{\color[HTML]{FD6864} \textbf{-9.3\%}}} & 41.53 & 38.98 & {\color[HTML]{FD6864} \textbf{-6.1\%}} \\
\multicolumn{1}{l|}{Llama-4-Maverick-17B-128E-Instruct} & 77.58 & 72.52 & \multicolumn{1}{l|}{{\color[HTML]{FD6864} \textbf{-7.0\%}}} & 68.10 & 57.15 & {\color[HTML]{FD6864} \textbf{-16.1\%}} \\
\multicolumn{1}{l|}{Llama-4-Scout-17B-16E-Instruct} & 67.49 & 45.03 & \multicolumn{1}{l|}{{\color[HTML]{FD6864} \textbf{-33.3\%}}} & 58.14 & 28.74 & {\color[HTML]{FD6864} \textbf{-50.6\%}} \\
\bottomrule
\end{tabular}%
}
\caption{Comparison results between English and Korean prompts of models whose main results are reported on English prompts. The \textit{diff} values are the relative difference in scores between two prompts. The specific prompts are detailed in Appendix~\ref{appendix:eval_prompts}.}
\label{tab:ko_prompt}
\end{table}

%% file: section/related_works.tex
\paragraph{Reliability Issues of Benchmarks}
Recent works~\cite{gema2025mmlu,vendrow2025largelanguagemodelbenchmarks} have raised concerns about the reliability of LLM benchmarks due to dataset noise and contamination. MMLU-Redux~\cite{gema2025mmlu} improved evaluation quality through systematic human reannotation, while GSM8K-Platinum~\cite{vendrow2025largelanguagemodelbenchmarks} refined arithmetic benchmarks via automated and manual error detection. MMLU-CF~\cite{zhao2024mmlucfcontaminationfreemultitasklanguage} prevents both unintentional and malicious contamination via sourcing diverse domains and question rewriting. LiveBench~\cite{white2025livebench} and LiveCodeBench~\cite{jain2025livecodebench} adopted dynamic evaluation protocols with temporal cutoffs to prevent future leakage.

\paragraph{Professional Benchmark}
With the rapid advancement of LLMs, more challenging benchmarks have become essential. GPQA~\cite{rein2024gpqa} and SuperGPQA~\cite{pteam2025supergpqascalingllmevaluation} assess graduate-level knowledge; MMLU-Pro~\cite{wang2024mmlupro} extends MMLU by increasing the share of college-level questions and expanding answer choices. Humanity’s Last Exam~\cite{phan2025humanitysexam} introduces a frontier benchmark composed of manually authored, research-level questions. 

\paragraph{Korean Benchmark}
While prior benchmarks focus primarily on English, recent efforts have produced Korean-specific evaluations~\cite{son-etal-2024-hae, kim-etal-2024-click}. Some rely on translated datasets~\cite{park-etal-2024-open, kim2025openkollmleaderboard2bridging, mmmlu, singh2024globalmmluunderstandingaddressing}, often with human post-editing, but these lack regional context, institutional norms, and domain-specific fluency. In contrast, native Korean benchmarks such as KMMLU~\cite{son2024kmmlumeasuringmassivemultitask}, KorMedMCQA~\cite{kweon2024kormedmcqamultichoicequestionanswering}, and KBL~\cite{kimyeeun-etal-2024-developing} address cultural and linguistic specificity. However, KMMLU~\cite{son2024kmmlumeasuringmassivemultitask} suffers from quality issues, including leaked answers, and KorMedMCQA~\cite{kweon2024kormedmcqamultichoicequestionanswering} and KBL~\cite{kimyeeun-etal-2024-developing} are limited to narrow domains. 

In this work, we introduce \redux{} and \pro{}, two contamination-free, industry grade benchmarks, providing a practical assessment of LLM capabilities in Korean industries.

%% file: section/conclusion.tex
We present two benchmarks constructed from real-world professional licensing exams, designed to reflect industrial domain knowledge and practical application standards. To ensure reliability, we identify and eliminate various sources of errors. Through extensive experiments, we evaluate the professional knowledge capabilities of LLMs across a wide range of domains. Our analysis further identifies key factors that influence performance, including region-specific knowledge, reasoning budget, and prompt language. We hope this work provides a foundation for more rigorous evaluation and continued advancement of real-world competence in language models.

%% file: section/appendix.tex
\section{Dataset Examples}\label{appendix:example_of_redux_and_pro}
\input{resources/tab_kmmlu-redux_pro_example}
The Table~\ref{tab:kmmlu-redux_pro_example} presents the example of~\redux and~\pro. While both benchmarks include domains such as management and accounting, \redux{} emphasizes industrial expert knowledge, whereas \pro{} focuses on practical professional expertise. This allows for a clear comparison of the different knowledge evaluations between the two benchmarks.

\section{Details of \redux~Errors}
\label{appendix:detail_redux_errors}
\subsection{Error Statistics}
We define a set of error types based on recurring issues observed during our analysis of the ~\original. Then, we find out the number of data instances per each error type as shown in Table~\ref{tab:redux_error_types_stats}. To identify leaked answer cases, we apply rule-based filtering using string-overlap heuristics. For other error types, including notation errors, bad clarity, and ill-posed questions, we leverage GPT-4o to assist in annotation.  Also, we provide the prompt used for annotation in Figure~\ref{fig:error_prompt}.

\begin{itemize}
    \item \textbf{Ill-posed Question} : The question lacks critical references or contextual information.
    \item \textbf{Leaked Answer} : The ground truth is explicitly stated within the question itself.
    \item \textbf{Notation Error} : Errors in mathematical expressions or chemical equations.
    \item \textbf{Bad Clarity} : The data itself is unclear and contains grammatical errors.
\end{itemize}

\input{resources/tab_redux_error_types_stats}
\input{resources/fig_error_annotation_prompt}

\subsection{Example of Error Types}
\label{appendix:example_error_types}

Table \ref{tab:example_error_types} provides examples of the error types in \ref{subsec:dataset_construction_redux} identified in \original. Each example illustrates a specific issue that affects benchmark reliability.

\section{Korean National Technical Qualification List of \redux}\label{appendix:redux_KNTQ_list}

\input{resources/fig_redux_source_distribution}

Table~\ref{tab:redux_detail_years_NTQ} presents the list of Korean National Technical Qualifications (KNTQs) included in our benchmark along with their corresponding official exam dates. To facilitate analysis, we categorize the 100 NTQs into Korean Standard Industrial Classification (KSIC) where mapping to the exam (see Figure~\ref{fig:redux_category}). This categorization enables structured evaluation across diverse domains and better reflects the real-world industrial fields.

\section{Details of \pro~Annotation}\label{appendix:annotation_guideline}
\subsection{Annotation Pipeline}
We first conduct OCR parsing with GPT-4o on PDF files of KNPL acquisition exams. With the parsed data, the main tasks for human annotators are: 1) reviewing parsing errors, 2) converting tables into latex format, and 3) converting images into text which conveys same meaning. If it is impossible to convert an image into text, we remove it. For the cases where multiple answers are allowed, commonly due to the ambiguity of question itself, we discard them.

As we leverage the official PDF files managed and controlled by the government, we can guarantee the correctness of answer label. This help us to save costs because we do not need experts for annotations nor the answer relabeling to avoid risk of data from online~\citep{gema2025mmlu,pteam2025supergpqascalingllmevaluation}. Before annotation, we explained the context of benchmark construction about the Korean professional license exams to human annotators. We present annotation instructions in Fig~\ref{fig:annotation_guidelines}. 

\input{resources/fig_annotation_guideline}

\subsection{Annotators Demographics}

\input{resources/tab_annotator_demo}

With 23 annotators, it took 8 business days to complete all annotations, including project setup. The detailed demographics of annotators are presented in Table~\ref{tab:annotator_demo}. The total amount of annotations was approximately \$14,000. The average hourly wage is 8.83 U.S. dollars, which is higher than the legal minimum wage at the time of hiring in South Korea.

\section{Evaluation Setup}\label{appendix:evaluation_details}

\input{resources/fig_eng_prompt}
\input{resources/fig_kor_prompt}

\subsection{Evaluation Prompts}\label{appendix:eval_prompts}
The figure~\ref{fig:en_prompt} and figure~\ref{fig:ko_prompt} present the prompt for the evaluation written in English and Korean, respectively. For the English prompt, we use the regex expression of \texttt{r"(?i)Answer[\^{}A-E]*: [\^{}A-E]*([A-E])"}. For the Korean prompt, we use \texttt{r"정답[\^{}A-E]*:[\^{}A-E]*([A-E])"}. The regex expressions for the flexible parsing are \texttt{r"Answer[\^{}A-E]*([A-E])|([A-E])\textbackslash)"} and \texttt{r"정답[\^{}A-E]*([A-E])|([A-E])\textbackslash)"}, respectively.

\subsection{Inference Engines}
Excluding closed models, the main inference engine we use is SGLang~\citep{zheng2024sglang}. However, for the Gemma 3 series and Mistral Small 3.1 Instruct models, we adopt vLLM~\citep{kwon2023efficient} due to their incompatibility with SGLang at the time of the paper writing.

\subsection{License Passing Criteria for \pro}\label{appendix:pro_scoring}
We follow the official scoring criteria used for each license examination. All licensing exams in \pro~are composed of multiple subjects. Candidates are typically required to score at least 40\% in every subject and achieve an average score of at least 60\%, except for the cases of the Certified Judicial Scrivener and the Lawyer. For the Certified Judicial Scrivener exam, candidates need only score at least 40\% in each subject, with no requirement regarding the average. The Lawyer license exam uses a relative grading system where only a certain proportion of top-scoring candidates pass. The usual cut-off point is approximately 54.22 (900 out of 1660), which we use as the passing threshold.

It is also important to note that our evaluation benchmark is text-based, not multi-modal. Therefore, we exclude questions that include images. In addition, candidates often need to go through multiple exam stages to obtain a license, with the later stages, such as the second or third, containing descriptive questions. However, since we only collect multiple-choice questions, the descriptive problems are excluded. Lastly, we exclude questions with multiple answers introduced by the ambiguity of the question.

\section{Detailed Results}

\subsection{The Results for Smaller (\textless 10B) Models}\label{appendix:main_result_smaller}

\input{resources/tab_main_results_appendix}

The Table~\ref{tab:main_results_smaller} presents the results of \pro~and \redux~for smaller (\textless 10B) models. Since many of the \textit{tiny} models in this table were used to construct \redux~through adversarial filtration, their \redux~scores are biased. Nevertheless, as shown by the results for larger models, models equipped with dense reasoning capabilities usually outperform their counterparts without reasoning (e.g., Qwen3-8B with and without ``thinking``).

\subsection{Breakdown of Results of \redux}\label{appendix:breakdown_redux}

\input{resources/tab_kmmlu_redux_category_result_full}

The Table~\ref{tab:redux_category_result_full} presents the breakdown results for 14 categories in \redux~across various LLMs.

\subsection{Breakdown of Results of \pro}\label{appendix:breakdown_pro}

\input{resources/tab_kmmlu_pro_category_result_full}

The Table~\ref{tab:pro_category_result_full} shows the breakdown results for all NPLs in \pro~across various LLMs. While the models relatively easily pass licensing in the medicine domain, they struggle in the Law and Tax\&Accounting domains.

\subsection{Performance Comparison Between Reasoning and Non-Reasoning Models}
\label{appendix:reasoning_vs_non_reasoning}
To further analyze the effects of \textit{reasoning} (or \textit{thinking}; we use these two terms interchangeably), we compare the scores of reasoning models with their corresponding non-reasoning models for each license in \pro{}. The Qwen3 series, EXAONE 4.0 series, and Claude 3.7 support both think-on/off modes, so we compare the scores under both modes. Other closed models, such as OpenAI's GPT-4.1 vs. O-series, are excluded since we do not have information whether the reasoning models share the same architecture with instruction-tuned models. Most of the scores are from Table~\ref{tab:pro_category_result_full}, except for DeepSeek-R1-Distill-Llama-70B~\citep{deepseekai2025deepseekr1incentivizingreasoningcapability}, which we evaluated newly. Table~\ref{tab:reasoning_vs_non_reasoning} presents the performance gains (\%) when reasoning is enabled.

\input{resources/tab_reasoning_vs_nonreasoning}

\input{resources/tab_redux_year_test_info}
\input{resources/tab_kmmlu_error_ex}

%% file: resources/tab_kmmlu-redux_pro_example.tex
\begin{table*}[!ht]
\centering
\resizebox{\textwidth}{!}{
\begin{tabular}{llp{7cm}p{7cm}}
\toprule
& & \redux & \pro \\
\midrule
\multirow{2}{*}{Question}
    &  & 전수검사가 불가능하여 반드시 샘플링검사를 하여야 하는 경우는? 
      & 액면주식의 주권을 발행한 비상장주식회사의 상법 제329조의2 소정의 주식분할에 관한 설명으로 옳은 것은? \\
    &  & \textit{\textcolor{gray}{Which case is necessary to conduct sampling inspections due to a complete inspection is not possible?}}
      & \textit{\textcolor{gray}{Which of the following statements about stock splits under Article 329-2 of the Commercial Act for unlisted stock companies that have issued share certificates is correct?}} \\ \midrule
\multirow{2}{*}{Options}
    &  & [ "전기제품의 출력전압의 측정", "주물제품의 내경가공에서 내경의 측정", "전구의 수입검사에서 전구의 점등시험", "진공관의 수입검사에서 진공관의 평균수명 추정"]
      & [ "주식분할을 하기 위해서는 주주총회의 특별결의를 거쳐야 한다.", "회사가 공고한 주권제출기간 중에 주주가 주권을 제출하면 그 시점에 주식분할의 효력이 발생한다.", "주식분할이 이루어져도 발행주식총수는 증가하지 않는다.", "주식분할이 이루어져도 1주의 액면금액은 감소하지 않는다.", "주식발행이 이루어지면 회사의 자본금이 증가한다." ] \\
    &  & \textit{\textcolor{gray}{["Measuring the output voltage of electrical products", "Measuring the inner diameter in machining of cast products", "Lighting test for imported light bulbs during inspection", "Estimating the average lifespan of vacuum tubes during import inspection"]}}
      & \textit{\textcolor{gray}{["A special resolution at the shareholders’ meeting is required to carry out a stock split.", "If a shareholder submits share certificates during the public notice period announced by the company, the stock split becomes effective at that time.", "The total number of issued shares does not increase even after a stock split.", "The face value per share does not decrease after a stock split.", "When shares are issued, the company’s capital increases."]}} \\ \midrule
\multirow{2}{*}{Answer}
    &  & 진공관의 수입검사에서 진공관의 평균수명 추정
      & 주식분할을 하기 위해서는 주주총회의 특별결의를 거쳐야 한다. \\
    &  & \textit{\textcolor{gray}{Estimating the average lifespan of vacuum tubes during import inspection}}
      & \textit{\textcolor{gray}{A special resolution at the shareholders’ meeting is required to carry out a stock split.}} \\ \midrule
\multirow{2}{*}{License name}
    &  & 품질경영기사 & 회계사 \\
    &  & \textit{\textcolor{gray}{Engineer Quality Management}} & \textit{\textcolor{gray}{Certified Public Accountant}} \\
\bottomrule
\end{tabular}
}
\caption{Examples of KMMLU-Redux and KMMLU-Pro. \textcolor{gray}{Gray} text represents English translations of the original Korean.}
\label{tab:kmmlu-redux_pro_example}
\end{table*}

%% file: resources/tab_redux_error_types_stats.tex
\begin{table}[!htp]
    \centering
    \begin{tabular}{c|c} \toprule 
    Error Type & \# of Questions (Ratio) \\ \midrule
    Ill-posed Question & 512 (1.46 \%) \\
    Leaked Answer & 42 (0.11\%) \\
    Notation Error & 846 (2.42\%) \\
    Bad Clarity & 1284 (3.67\%) \\ \bottomrule
    \end{tabular}
    \caption{Statistics of the error types}
    \label{tab:redux_error_types_stats}
\end{table}

%% file: resources/fig_error_annotation_prompt.tex
\begin{figure*}[t]
\centering
\begin{tcolorbox}[
  title=GPT-4o Error Annotation Prompt, 
  colframe=black!80!white,   
  colback=gray!10,          
  coltitle=white,           
  colbacktitle=black!80!white, 
  fonttitle=\bfseries,      
  rounded corners,          
  boxsep=3pt,               
  width=\textwidth          
]

You are a helpful assistant that annotates error types in questions. \\
- Ill-posed question stands for which question miss the critical reference information to solve the question (such as table, formular, image, etc.). \\
- Notation correct stands for which question has correct math and chemical notation. Criteria of correctness whether the notation is impact to solve the question. (e.g. m2 is incorrect, but m\^2 is correct. 센티미터 is correct.) \\
- Grammar correct stands for which question has correct grammar and spelling. Criteria of correctness whether the notation is impact to understand the question. (e.g. 상 담 기 법 보 다 는 상 담 자 의 인 간 적 자 질 과 진 솔 한 태 도 를 중 시 한 다 . is grammatically incorrect due to spacing error.) \\ \\

Please check the following question whether it has error types. \\
Check ill-posed question True/False. \\
Check notation correct True/False. \\
Check grammar correct True/False. \\
Then, if there is any error, please explain the reason. \\
\\
\*\* Question \*\*
\\
\{\{\textsl{question}\}\} \\
\\
\end{tcolorbox}
\caption{The prompt is used for error type annotation. Each sample is annotated as an error if the respective field returns True. The 'Grammar Correct' field is used to detect 'Bad Clarity' cases.}
\label{fig:error_prompt}
\end{figure*}

%% file: resources/fig_redux_source_distribution.tex
\begin{figure}
    \centering
    \includegraphics[width=\linewidth]{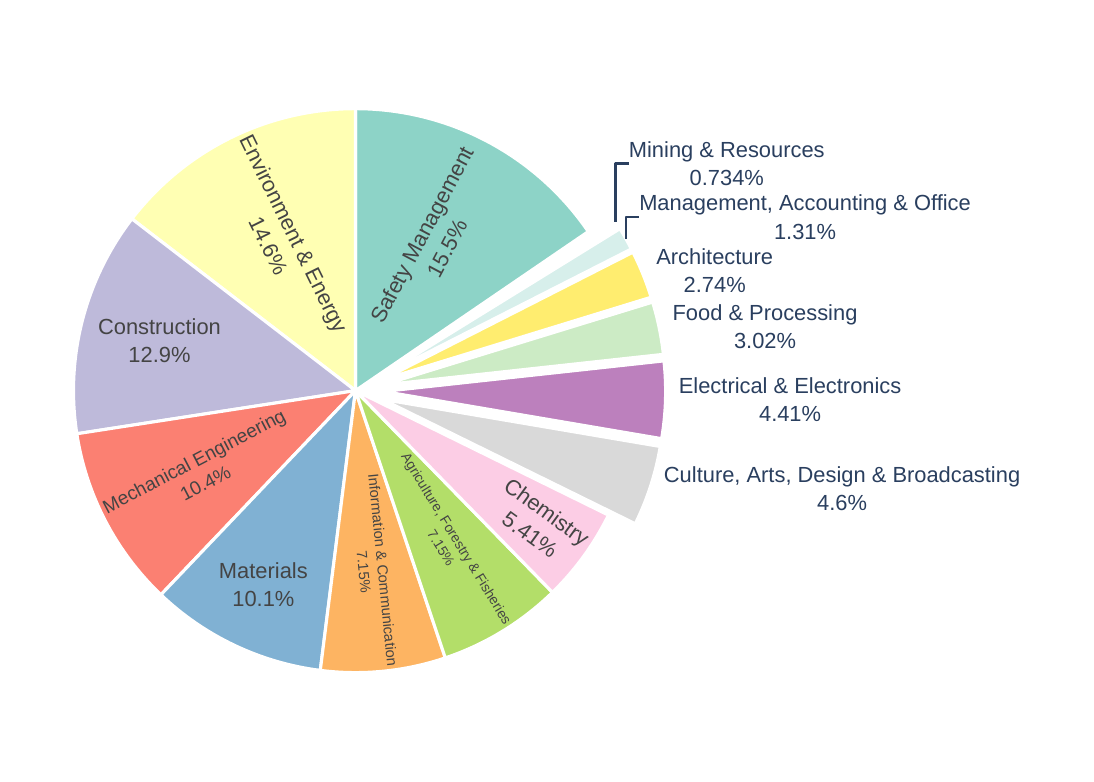}
    \caption{Domain distribution of problems in \redux. The total size of the dataset is 2,587.}
    \label{fig:redux_category}
\end{figure}

%% file: resources/fig_annotation_guideline.tex
\begin{figure*}[!tbp]
 \centering
  \resizebox{\textwidth}{!}{
  
  \begin{minipage}[c]{0.5\linewidth} 
    \centering
    \includegraphics[width=\linewidth]{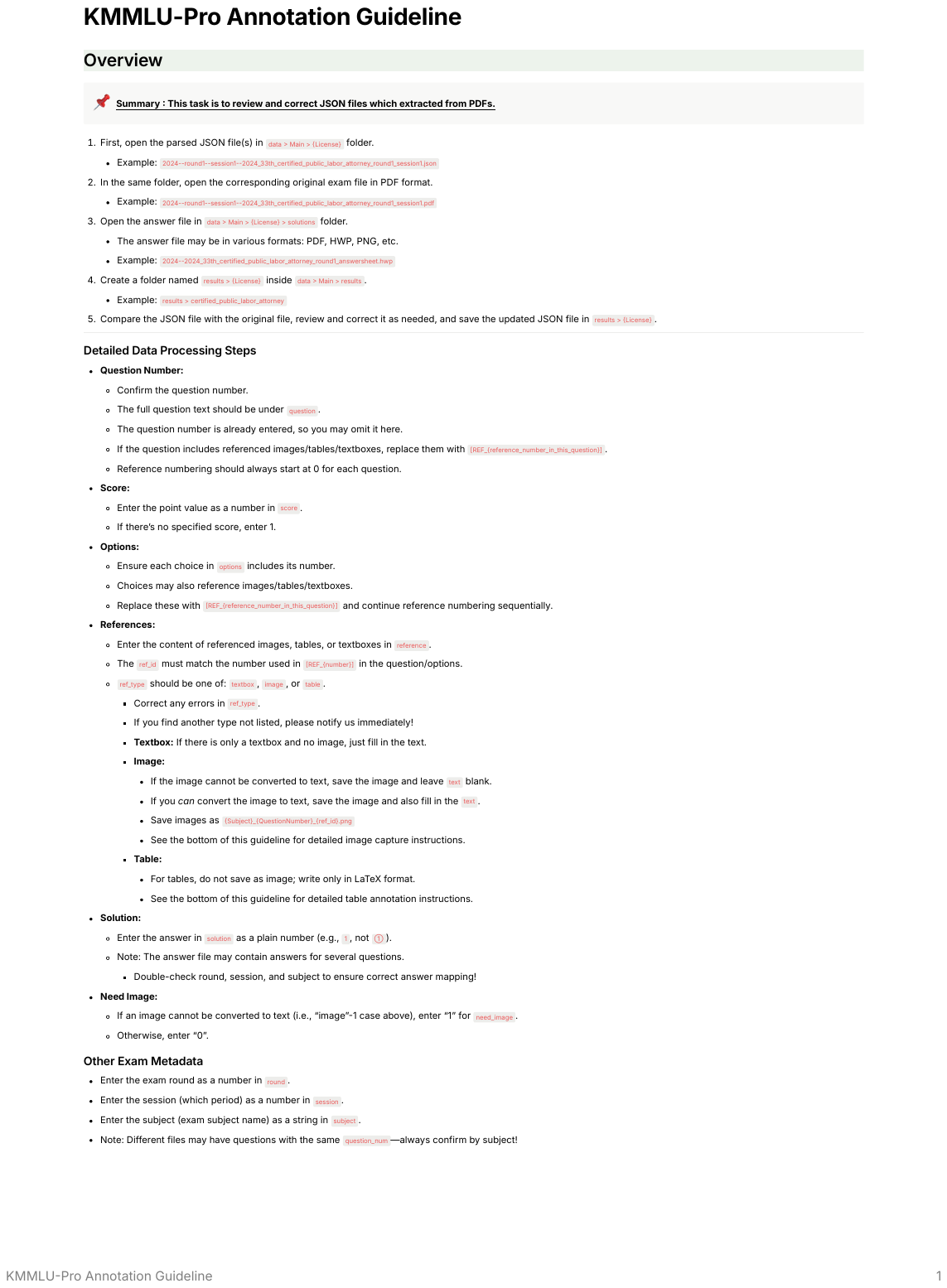}
  \end{minipage}%
  \hspace{\dimexpr0.01\linewidth}
  \begin{minipage}[c]{0.5\linewidth}  
    \centering
    \includegraphics[width=\linewidth]{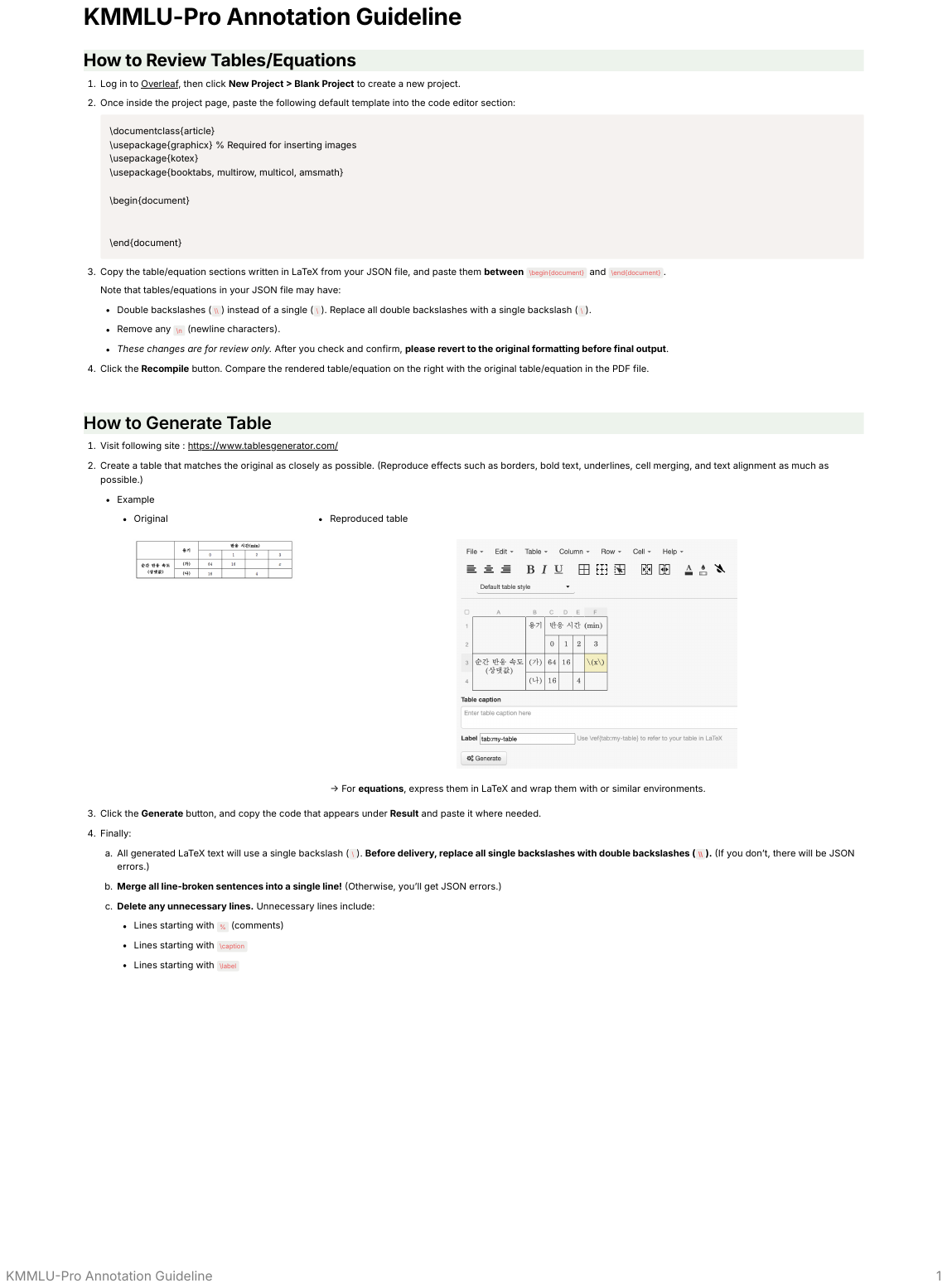}
  \end{minipage}
}
  \caption{Excerpt from the translated annotation guidelines for converting PDF documents into structured text. We carefully instruct LaTeX table formatting.}
  \label{fig:annotation_guidelines}
\end{figure*}

%% file: resources/tab_annotator_demo.tex
\begin{table}[h]
\centering
\scriptsize
\begin{tabular}{@{}lll@{}}
\toprule
Category & Demographics & Counts \\ \midrule
\multirow{2}{*}{Gender} & Female & 23 \\
 & Male & - \\ \midrule
\multirow{3}{*}{Age} & 20s & 13 \\
 & 30s & 8 \\
 & 40s & 2 \\ \midrule
\multirow{2}{*}{Academic background} & Bachelor's degrees & 20 \\
 & Master's degrees & 3 \\ \bottomrule
\end{tabular}%

\caption{Demographics of the human annotators for \pro.}
\label{tab:annotator_demo}
\end{table}

%% file: resources/fig_eng_prompt.tex
\begin{figure*}[!h]
\centering
\begin{tcolorbox}[
  title=English Prompt, 
  colframe=black!80!white,   
  colback=gray!10,          
  coltitle=white,           
  colbacktitle=black!80!white, 
  fonttitle=\bfseries,      
  rounded corners,          
  boxsep=3pt,               
  width=\textwidth          
]

Answer the following multiple choice question. The last line of your response should be of the following format: 'Answer: \$LETTER' (without quotes) where LETTER is one of ABCD. Think step by step before answering. \\
\\
\{\{\textsl{question}\}\} \\
\\
A) \{\{\textsl{option A}\}\} \\
B) \{\{\textsl{option B}\}\} \\
C) \{\{\textsl{option C}\}\} \\
D) \{\{\textsl{option D}\}\} \\

\end{tcolorbox}
\caption{The English prompt used for evaluating LLMs on our \redux~and \pro. This prompt is exactly same with the prompt used for Multiple-Choices Question Answering (MCQA) in OpenAI's simple-evals repository. The number of options is adjusted for each problems.}
\label{fig:en_prompt}
\end{figure*}

%% file: resources/fig_kor_prompt.tex
\begin{figure*}[!th]
\centering
\begin{tcolorbox}[
  title=Korean Prompt, 
  colframe=black!80!white,   
  colback=gray!10,          
  coltitle=white,           
  colbacktitle=black!80!white, 
  fonttitle=\bfseries,      
  rounded corners,          
  boxsep=3pt,               
  width=\textwidth          
]

다음 문제에 대해 정답을 고르세요. 당신의 최종 정답은 ABCD 중 하나이고, "정답:" 뒤에 와야 합니다. 정답을 고르기 전에 차근차근 생각하고 추론하세요. \\
\\
\{\{\textsl{question}\}\} \\
\\
A) \{\{\textsl{option A}\}\} \\
B) \{\{\textsl{option B}\}\} \\
C) \{\{\textsl{option C}\}\} \\
D) \{\{\textsl{option D}\}\} \\

\end{tcolorbox}
\caption{The Korean prompt used for evaluating LLMs on our \redux~and \pro. This prompt is translated version of the English prompt. The number of options is adjusted for each problems.}
\label{fig:ko_prompt}
\end{figure*}

%% file: resources/tab_main_results_appendix.tex
\begin{table*}[!ht]
\centering
\footnotesize
\resizebox{\textwidth}{!}{
\begin{tabular}{@{}lcccc@{}}
\toprule
\multicolumn{1}{l|}{} & KMMLU-Redux & \multicolumn{2}{c|}{KMMLU-Pro} &  \\
\multicolumn{1}{l|}{} & Acc & Acc & \multicolumn{1}{c|}{
\# of passed KNPLs} & Avg. Acc (micro) \\
\midrule
\multicolumn{1}{r|}{ DeepSeek-R1-Distill-Qwen-1.5B~\citep{deepseekai2025deepseekr1incentivizingreasoningcapability}} & 21.30$^\star$ & 20.55 & \multicolumn{1}{c|}{0/14} & 20.91 \\
\multicolumn{1}{r|}{ Llama 3.2 3B Instruct~\citep{llama3.2}} & 17.59$^\star$ & 25.53 & \multicolumn{1}{c|}{0/14} & 21.73 \\
\multicolumn{1}{r|}{ Gemma 3 4B IT~\citep{gemma3}} & 25.09$^\star$ & 32.86 & \multicolumn{1}{c|}{0/14} & 29.14 \\
\multicolumn{1}{r|}{ Qwen 2.5 3B Instruct~\citep{qwen2025qwen25technicalreport}} & 24.74$^\star$ & 33.27 & \multicolumn{1}{c|}{0/14} & 29.19 \\
\multicolumn{1}{r|}{ Qwen3-1.7B~\citep{yang2025qwen3technicalreport}} & 28.99 & 30.42 & \multicolumn{1}{c|}{0/14} & 29.74 \\
\multicolumn{1}{r|}{ Kanana Nano 2.1B Instruct~\citep{kananallmteam2025kananacomputeefficientbilinguallanguage}} & 27.25$^\star$ & 32.60 & \multicolumn{1}{c|}{0/14} & 30.04 \\
\multicolumn{1}{r|}{ Aya Expanse 8B~\citep{dang2024ayaexpansecombiningresearch}} & 28.30 & 31.65 & \multicolumn{1}{c|}{0/14} & 30.05 \\
\multicolumn{1}{r|}{ HyperCLOVAX-SEED-Text-Instruct-1.5B~\citep{hcxseed}} & 33.94 & 30.13 & \multicolumn{1}{c|}{0/14} & 31.95 \\
\multicolumn{1}{r|}{ Llama 3.1 8B Instruct~\citep{llama3.1}} & 31.89 & 33.81 & \multicolumn{1}{c|}{0/14} & 32.89 \\
\rowcolor[rgb]{0.9,0.9,0.9} \multicolumn{1}{r|}{Qwen3-1.7B~{\smaller[2]~(w/ thinking)}~\citep{yang2025qwen3technicalreport}} & 37.80 & 38.27 & \multicolumn{1}{c|}{1/14} & 38.05 \\
\multicolumn{1}{r|}{ EXAONE 4.0 1.2B~\citep{research2025exaone40unifiedlarge}} & 40.43 & 37.48 &  \multicolumn{1}{c|}{0/14} & 38.89 \\
\rowcolor[rgb]{0.9,0.9,0.9} \multicolumn{1}{r|}{Ko-R1-7B-v2.1~\citep{ko-r1-7b-v2.1}} & 41.94 & 38.70 & \multicolumn{1}{c|}{1/14} & 40.25 \\
 \rowcolor[rgb]{0.9,0.9,0.9} \multicolumn{1}{r|}{ EXAONE 4.0 1.2B~{\smaller[2]~(w/ thinking)}~\citep{research2025exaone40unifiedlarge}} & 46.85 & 42.69 &  \multicolumn{1}{c|}{0/14} & 44.68 \\
\multicolumn{1}{r|}{ Qwen3-8B~\citep{yang2025qwen3technicalreport}} & 49.25 & 46.92 & \multicolumn{1}{c|}{1/14} & 48.03 \\
\rowcolor[rgb]{0.9,0.9,0.9} \multicolumn{1}{r|}{Qwen3-8B~{\smaller[2]~(w/ thinking)}~\citep{yang2025qwen3technicalreport}} & 58.79 & 55.27 & \multicolumn{1}{c|}{3/14} & 56.95 \\
\bottomrule
\end{tabular}
}
\caption{The main evaluation results of \redux~and \pro~benchmarks on smaller (\textless~10B) LLMs. \colorbox{gray!20}{The gray-shaded models} are the dense-reasoning models. The \redux~scores with $^\star$ are biased as these models are used for the dataset filtration (Section~\ref{subsubsec:challenging_exams}).}
\label{tab:main_results_smaller}
\end{table*}

%% file: resources/tab_kmmlu_redux_category_result_full.tex
\begin{table*}[!ht]
\centering
\resizebox{\textwidth}{!}{
\begin{tabular}{@{}cr|cccccccccccccc|c@{}}
\toprule
 \multicolumn{2}{l|}{Domain}   & \rotatebox{80}{Safety Management} & \rotatebox{80}{Environment \& Energy} & \rotatebox{80}{Construction} & \rotatebox{80}{Mechanical Engineering} & \rotatebox{80}{Materials}  & \rotatebox{80}{Information \& Communication} & \rotatebox{80}{Agriculture, Forestry \& Fisheries} & \rotatebox{80}{Chemistry} & \rotatebox{80}{Culture, Arts, Design \& Broadcasting} & \rotatebox{80}{Electrical \& Electronics} & \rotatebox{80}{Food \& Processing} & \rotatebox{80}{Architecture} & \rotatebox{80}{Management, Accounting \& Office} & \rotatebox{80}{Mining \& Resources} & \rotatebox{80}{Avg.} \\ \midrule
\multicolumn{17}{l}{Open-weight Models} \\ \midrule
\rowcolor[rgb]{0.9,0.9,0.9} \cellcolor[rgb]{1.0,1.0,1.0}\textless~5B &  DeepSeek-R1-Distill-Qwen-1.5B & 20.50 & 17.77 & 24.32 & 20.00 & 24.43 & 26.49 & 17.84 & 20.71 & 19.33 & 20.18 & 15.38 & 18.31 & 20.59 & 31.58 & 21.30$^\star$  \\
 & Llama 3.2 3B Instruct & 13.00 & 12.20 & 13.81 & 14.81 & 11.07 & 15.68 & 16.76 & 11.43 & 17.65 & 7.89 & 16.67 & 11.27 & 5.88 & 15.79 & 17.59$^\star$  \\
 & Qwen 2.5 3B Instruct & 16.50 & 17.24 & 15.32 & 17.41 & 15.27 & 20.54 & 18.38 & 11.43 & 26.05 & 16.67 & 11.54 & 15.49 & 14.71 & 10.53 & 24.74$^\star$  \\
 & Gemma 3 4B IT & 23.00 & 23.87 & 22.52 & 25.56 & 20.23 & 28.65 & 32.43 & 25.71 & 21.01 & 27.19 & 21.79 & 21.13 & 14.71 & 10.53 & 25.09$^\star$  \\
 & Kanana Nano 2.1B Instruct & 24.75 & 27.32 & 25.83 & 26.67 & 29.77 & 22.16 & 26.49 & 23.57 & 25.21 & 28.07 & 28.21 & 28.17 & 26.47 & 10.53 & 27.25$^\star$  \\
& Qwen3-1.7B & 30.50 & 27.06 & 27.93 & 36.67 & 32.82 & 29.73 & 23.24 & 30.00 & 23.53 & 28.95 & 25.64 & 23.94 & 20.59 & 15.79 & 28.99 \\
 & HyperCLOVAX-SEED-Text-Instruct-1.5B & 28.00 & 25.99 & 30.93 & 32.59 & 33.21 & 29.19 & 32.97 & 22.86 & 35.29 & 28.07 & 32.05 & 32.39 & 17.65 & 26.32 & 33.94 \\
\rowcolor[rgb]{0.9,0.9,0.9} \cellcolor[rgb]{1.0,1.0,1.0} & Qwen3-1.7B~{\smaller[2]~(w/ thinking)} & 37.25 & 35.54 & 33.03 & 39.63 & 40.08 & 41.62 & 36.76 & 40.71 & 35.29 & 46.49 & 42.31 & 29.58 & 41.18 & 42.11 & 37.80 \\
 & EXAONE 4.0 1.2B & 44.00 & 37.14 & 39.34 & 44.07 & 44.27 & 48.11 & 38.38 & 34.29 & 41.18 & 35.96 & 32.05 & 38.03 & 29.41 & 21.05 & 40.43 \\ 
  \rowcolor[rgb]{0.9,0.9,0.9} \cellcolor[rgb]{1.0,1.0,1.0} & EXAONE 4.0 1.2B~{\smaller[2]~(w/ thinking)} & 47.50 & 44.03 & 49.30 & 47.41 & 50.76 & 50.81 & 42.70 & 50.00 & 44.54 & 50.00 & 38.46 & 49.30 & 32.35 &  21.05 & 46.85 \\ \midrule
\textless~10B  & Aya Expanse 8B & 25.00 & 20.42 & 23.72 & 21.11 & 24.05 & 29.19 & 29.19 & 21.43 & 23.53 & 22.81 & 20.51 & 30.99 & 26.47 & 21.05 & 28.30 \\
 & Llama 3.1 8B Instruct & 30.50 & 25.99 & 23.12 & 28.15 & 30.53 & 33.51 & 32.97 & 22.14 & 35.29 & 24.56 & 29.49 & 26.76 & 29.41 & 31.58 & 31.89 \\
 \rowcolor[rgb]{0.9,0.9,0.9} \cellcolor[rgb]{1.0,1.0,1.0} & Ko-R1-7B-v2.10 & 42.50 & 37.93 & 39.94 & 41.48 & 38.55 & 50.81 & 37.30 & 52.14 & 49.58 & 50.00 & 33.33 & 33.80 & 50.00 & 36.84 & 41.94 \\
 & Qwen3-8B & 47.00 & 48.01 & 43.84 & 49.63 & 56.11 & 56.22 & 41.62 & 58.57 & 47.06 & 60.53 & 46.15 & 43.66 & 50.00 & 31.58 & 49.25 \\
\rowcolor[rgb]{0.9,0.9,0.9} \cellcolor[rgb]{1.0,1.0,1.0} & Qwen3-8B~{\smaller[2]~(w/ thinking)} & 55.75 & 57.82 & 53.15 & 67.78 & 64.89 & 62.16 & 45.95 & 65.71 & 57.14 & 66.67 & 56.41 & 53.52 & 58.82 & 63.16 & 58.79 \\ \midrule
\textless~20B & Gemma 3 12B IT & 44.50 & 42.18 & 46.25 & 44.81 & 46.95 & 50.81 & 51.35 & 50.71 & 57.98 & 48.25 & 46.15 & 39.44 & 50.00 & 42.11 & 46.70 \\
 & Phi-4 (14B) & 48.50 & 47.75 & 48.05 & 52.59 & 49.62 & 52.43 & 44.86 & 58.57 & 54.62 & 56.14 & 50.00 & 36.62 & 52.94 & 36.84 & 49.75 \\
  & Qwen3-14B & 52.75 & 55.70 & 51.95 & 63.33 & 61.45 & 65.95 & 47.57 & 61.43 & 57.98 & 67.54 & 52.56 & 59.15 & 61.76 & 47.37 & 57.25 \\
  \rowcolor[rgb]{0.9,0.9,0.9} \cellcolor[rgb]{1.0,1.0,1.0} & Qwen3-14B~{\smaller[2]~(w/ thinking)} & 64.50 & 66.58 & 60.36 & 70.00 & 72.90 & 68.65 & 54.05 & 73.57 & 63.87 & 75.44 & 51.28 & 57.75 & 70.59 & 68.42 & 65.71 \\\midrule
\textless~32B & Aya Expanse 32B & 30.50 & 32.10 & 27.03 & 38.15 & 35.50 & 39.46 & 37.84 & 28.57 & 33.61 & 35.09 & 32.05 & 35.21 & 26.47 & 21.05 & 33.05 \\
 & Mistral Small 3.1 Instruct (24B) & 45.00 & 50.40 & 51.95 & 55.93 & 59.54 & 58.38 & 51.35 & 58.57 & 56.30 & 58.77 & 47.44 & 52.11 & 58.82 & 31.58 & 52.92 \\
& Gemma 3 27B IT & 49.50 & 51.19 & 47.45 & 57.41 & 58.02 & 62.70 & 49.73 & 60.71 & 59.66 & 62.28 & 56.41 & 47.89 & 67.65 & 31.58 & 54.04 \\
 & Qwen3-30B-A3B & 54.25 & 57.56 & 54.65 & 58.52 & 64.50 & 61.08 & 51.89 & 65.00 & 62.18 & 68.42 & 58.97 & 56.34 & 58.82 & 52.63 & 58.41 \\
  & EXAONE 4.0 32B & 61.00 & 62.86 & 61.26 & 70.74 & 74.05 & 70.81 & 60.54 & 67.86 & 74.79 & 63.16 & 55.13 & 49.30 & 61.76 & 42.11 & 64.79 \\ 
 & Qwen3-32B & 59.50 & 64.19 & 60.66 & 68.52 & 72.52 & 69.19 & 58.38 & 68.57 & 74.79 & 72.81 & 58.97 & 53.52 & 70.59 & 63.16 & 64.98 \\
 \rowcolor[rgb]{0.9,0.9,0.9} \cellcolor[rgb]{1.0,1.0,1.0} & Qwen3-30B-A3B~{\smaller[2]~(w/ thinking)} & 63.00 & 64.19 & 62.46 & 69.63 & 70.23 & 69.19 & 56.76 & 69.29 & 68.91 & 72.81 & 53.85 & 56.34 & 70.59 & 68.42 & 65.25 \\
 \rowcolor[rgb]{0.9,0.9,0.9} \cellcolor[rgb]{1.0,1.0,1.0} & QwQ 32B & 61.25 & 67.64 & 68.17 & 71.11 & 74.05 & 72.97 & 58.38 & 71.43 & 72.27 & 71.05 & 53.85 & 56.34 & 79.41 & 52.63 & 67.34 \\
 \rowcolor[rgb]{0.9,0.9,0.9} \cellcolor[rgb]{1.0,1.0,1.0} & Qwen3-32B~{\smaller[2]~(w/ thinking)} & 64.50 & 68.97 & 66.97 & 74.07 & 75.57 & 70.81 & 62.70 & 74.29 & 68.91 & 72.81 & 64.10 & 53.52 & 73.53 & 57.89 & 68.77 \\ 
   \rowcolor[rgb]{0.9,0.9,0.9} \cellcolor[rgb]{1.0,1.0,1.0} & EXAONE 4.0 32B~{\smaller[2]~(w/ thinking)} & 70.50 & 71.35 & 68.17 & 74.44 & 83.21 & 76.76 & 68.65 & 74.29 & 77.31 & 74.44 & 69.23 & 57.75 & 82.35 & 52.63 & 72.71 \\ \midrule
\textgreater~70B & Llama 3.3 70B Instruct & 52.25 & 54.11 & 52.85 & 62.59 & 59.92 & 64.32 & 52.43 & 50.71 & 60.50 & 60.53 & 55.13 & 53.52 & 64.71 & 36.84 & 56.17 \\
 & C4AI Command A (111B) & 56.75 & 66.58 & 63.06 & 63.33 & 64.89 & 67.57 & 62.16 & 60.71 & 68.91 & 64.04 & 62.82 & 59.15 & 55.88 & 47.37 & 62.93 \\
 & DeepSeek V3 (671B) & 62.50 & 62.33 & 63.66 & 66.30 & 72.52 & 72.97 & 62.70 & 66.43 & 67.23 & 69.30 & 69.23 & 59.15 & 61.76 & 63.16 & 65.64 \\
  & Llama-4-Scout-17B-16E-Instruct & 64.00 & 67.64 & 65.77 & 69.63 & 77.10 & 71.35 & 61.08 & 69.29 & 66.39 & 71.93 & 62.82 & 57.75 & 64.71 & 57.89 & 67.49 \\
 & Qwen3-235B-A22B & 64.25 & 72.68 & 66.37 & 71.11 & 82.06 & 72.43 & 65.41 & 71.43 & 68.07 & 73.68 & 65.38 & 52.11 & 67.65 & 47.37 & 69.54 \\
 \rowcolor[rgb]{0.9,0.9,0.9} \cellcolor[rgb]{1.0,1.0,1.0} & Qwen3-235B-A22B~{\smaller[2]~(w/ thinking)} & 69.75 & 75.86 & 71.47 & 81.11 & 85.11 & 76.22 & 64.86 & 75.00 & 70.59 & 78.95 & 73.08 & 61.97 & 82.35 & 68.42 & 74.49 \\
  & Llama-4-Maverick-17B-128E-Instruct & 73.50 & 76.13 & 78.38 & 79.26 & 83.97 & 80.54 & 77.30 & 75.00 & 78.15 & 79.82 & 73.08 & 71.83 & 82.35 & 73.68 & 77.58 \\
 \rowcolor[rgb]{0.9,0.9,0.9} \cellcolor[rgb]{1.0,1.0,1.0} & DeepSeek R1 (671B) & 72.50 & 76.39 & 79.58 & 80.74 & 85.11 & 81.62 & 82.70 & 77.14 & 79.83 & 81.58 & 75.64 & 63.38 & 85.29 & 73.68 & 78.51 \\ \midrule
\multicolumn{17}{l}{Closed Models} \\ \midrule
 & GPT-4.1 mini~{\smaller[2]~(2024-04-14)} & 61.00 & 64.99 & 63.66 & 69.26 & 74.81 & 72.43 & 63.78 & 73.57 & 70.59 & 66.67 & 75.64 & 56.34 & 73.53 & 57.89 & 67.03 \\
 \rowcolor[rgb]{0.9,0.9,0.9} \cellcolor[rgb]{1.0,1.0,1.0} &  o3-mini~{\smaller[2]~(2025-01-31)} & 63.75 & 66.05 & 63.96 & 72.96 & 73.66 & 75.14 & 63.78 & 69.29 & 72.27 & 75.44 & 66.67 & 46.48 & 76.47 & 57.89 & 67.84 \\
 \rowcolor[rgb]{0.9,0.9,0.9} \cellcolor[rgb]{1.0,1.0,1.0} & Grok-3-mini-beta & 69.75 & 69.76 & 69.07 & 73.33 & 83.97 & 77.30 & 65.95 & 70.71 & 68.07 & 73.68 & 61.54 & 59.15 & 76.47 & 73.68 & 71.47 \\
 & Grok-3-beta & 70.00 & 70.56 & 68.47 & 76.30 & 83.97 & 76.76 & 69.19 & 75.71 & 74.79 & 69.30 & 70.51 & 71.83 & 73.53 & 57.89 & 72.90 \\
 \rowcolor[rgb]{0.9,0.9,0.9} \cellcolor[rgb]{1.0,1.0,1.0} & o4-mini~{\smaller[2]~(2025-04-16)} & 67.75 & 73.74 & 76.28 & 78.52 & 82.06 & 81.08 & 76.76 & 78.57 & 80.67 & 79.82 & 71.79 & 61.97 & 79.41 & 78.95 & 75.80 \\
 & GPT-4.1 & 71.50 & 74.27 & 72.07 & 77.41 & 85.11 & 81.62 & 80.00 & 74.65 & 76.47 & 77.19 & 75.64 & 61.97 & 76.47 & 68.42 & 75.86 \\
 & Claude 3.7 Sonnet & 70.75 & 76.92 & 76.28 & 80.37 & 83.97 & 77.84 & 78.92 & 78.57 & 76.47 & 78.07 & 75.64 & 64.79 & 79.41 & 68.42 & 76.88 \\
 \rowcolor[rgb]{0.9,0.9,0.9} \cellcolor[rgb]{1.0,1.0,1.0} & Claude 3.7 Sonnet~{\smaller[2]~(w/ thinking)} & 71.50 & 80.37 & 79.88 & 80.74 & 87.02 & 82.16 & 80.00 & 82.86 & 80.67 & 79.82 & 73.08 & 70.42 & 85.29 & 68.42 & 79.36 \\
 \rowcolor[rgb]{0.9,0.9,0.9} \cellcolor[rgb]{1.0,1.0,1.0} & o3 & 77.75 & 78.25 & 77.78 & 80.37 & 85.50 & 78.92 & 83.24 & 78.17 & 84.03 & 85.09 & 82.05 & 66.20 & 85.29 & 78.95 & 79.92 \\
 \rowcolor[rgb]{0.9,0.9,0.9} \cellcolor[rgb]{1.0,1.0,1.0} & o1 & 75.75 & 79.58 & 81.98 & 81.11 & 90.84 & 80.54 & 84.86 & 77.86 & 84.03 & 81.58 & 83.33 & 70.42 & 85.29 & 73.68 & 81.14 \\ \bottomrule
\end{tabular}
}
\caption{The breakdown results for 14 categories in \redux. \colorbox{gray!20}{The gray-shaded models} are the dense-reasoning models. The scores with $^\star$ are biased as these models are used for the dataset filtration (Section~\ref{subsubsec:challenging_exams}).} 
\label{tab:redux_category_result_full}
\end{table*}

%% file: resources/tab_kmmlu_pro_category_result_full.tex
\begin{table*}[!ht]
\centering
\resizebox{\textwidth}{!}{
\begin{tabular}{@{}cr|cccc|ccc|cc|ccccc|cc@{}}
\toprule
 \multicolumn{2}{l|}{Domain} & \multicolumn{4}{c|}{Law} & \multicolumn{3}{c|}{Tax \& Accounting} & \multicolumn{2}{c|}{Value Estimation} & \multicolumn{5}{c|}{Medical} & & \\ \midrule
 \multicolumn{2}{l|}{Names of KNPLs}   & \rotatebox{80}{Judicial Scrivener} & \rotatebox{80}{Lawyer} & \rotatebox{80}{Public Labor Attorney} & \rotatebox{80}{Patent Attorney} & \rotatebox{80}{Public Accountant (CPA)}  & \rotatebox{80}{Tax Accountant} & \rotatebox{80}{Customs Broker} & \rotatebox{80}{Damage Adjuster (CDA)} & \rotatebox{80}{Appraiser} & \rotatebox{80}{Doctor of Korean Medicine} & \rotatebox{80}{Dentist} & \rotatebox{80}{Pharmacist} & \rotatebox{80}{Herb Pharmacist} & \rotatebox{80}{Physician} & \rotatebox{80}{Avg.} & \rotatebox{80}{\# of passed NPLs} \\ \midrule
\multicolumn{17}{l}{Open-weight Models} \\ \midrule
\rowcolor[rgb]{0.9,0.9,0.9} \cellcolor[rgb]{1.0,1.0,1.0} \textless~5B & DeepSeek-R1-Distill-Qwen-1.5B & 15.66 & 13.33 & 24.27 & 25.69 & 23.56 & 18.07 & 18.24 & 20.00 & 21.94 & 20.49 & 18.49 & 23.25 & 23.77 & 18.67 & 20.55 & 0/14 \\
 & Llama 3.2 3B Instruct & 27.27 & 24.67 & 25.94 & 20.18 & 16.35 & 23.95 & 23.9 & 23.33 & 19.39 & 23.26 & 26.45 & 36.53 & 30.74 & 28.67 & 25.53 & 0/14 \\
 & HyperCLOVAX-SEED-Text-Instruct-1.5B & 24.24 & 24.67 & 30.13 & 26.61 & 25.48 & 29.83 & 30.19 & 41.67 & 25.00 & 30.90 & 27.31 & 36.16 & 35.66 & 33.33& 30.13 & 0/14 \\
  & Qwen3-1.7B & 19.19 & 20.67 & 30.96 & 32.11 & 21.63 & 21.01 & 23.9 & 30.0 & 29.08 & 30.56 & 30.32 & 46.49 & 46.72 & 33.33 & 30.42 & 0/14 \\
 & Kanana Nano 2.1B Instruct & 19.19 & 24.67 & 30.13 & 22.02 & 23.08 & 26.47 & 26.42 & 28.33 & 26.53 & 42.01 & 31.61 & 52.40 & 44.67 & 38.67 & 32.60 & 0/14 \\
 & Gemma 3 4B IT & 23.23 & 26.67 & 32.22 & 26.61 & 27.4 & 26.47 & 27.67 & 29.17 & 23.47 & 37.15 & 36.56 & 49.08 & 43.03 & 36.00 & 32.86 & 0/14 \\
 & Qwen 2.5 3B Instruct & 20.20 & 25.33 & 35.15 & 24.77 & 21.63 & 23.11 & 30.19 & 35.00 & 27.04 & 37.85 & 35.05 & 47.97 & 52.46 & 34.67 & 33.27 & 0/14 \\
 & EXAONE 4.0 1.2B~{\smaller[2]~(w/ thinking)} & 26.26 & 22.00 & 35.98 & 25.69 & 30.77 & 41.60 & 32.70 & 44.17 & 37.24 & 41.80 & 37.63 & 54.24 & 41.80 & 44.67 & 37.48 & 0/14 \\ 
\rowcolor[rgb]{0.9,0.9,0.9} \cellcolor[rgb]{1.0,1.0,1.0} & Qwen3-1.7B~{\smaller[2]~(w/ thinking)} & 25.25 & 25.33 & 31.8 & 34.86 & 25.48 & 25.63 & 29.56 & 34.17 & 36.22 & 44.79 & 38.06 & {\color[HTML]{3166FF} \textbf{61.99}} & 59.43 & 44.67 & 38.27 & 1/14 \\ 
\rowcolor[rgb]{0.9,0.9,0.9} \cellcolor[rgb]{1.0,1.0,1.0} & EXAONE 4.0 1.2B~{\smaller[2]~(w/ thinking)} & 28.29 & 26.00 & 39.75 & 37.61 & 38.94 & 36.55 & 43.40 & 53.33 & 43.88 & 42.71 & 39.35 & 63.47 & 47.13 & 50.67 & 42.69 & 0/14 \\ \midrule
\textless~10B  & Aya Expanse 8B & 20.20 & 20.00 & 30.54 & 22.02 & 23.56 & 18.49 & 37.11 & 33.33 & 22.45 & 36.11 & 29.25 & 50.18 & 46.31 & 42.00 & 31.65 & 0/14 \\
 & Llama 3.1 8B Instruct & 28.28 & 21.33 & 30.96 & 23.85 & 21.15 & 23.53 & 33.33 & 33.33 & 26.02 & 35.42 & 33.76 & 54.61 & 50.82 & 42.00 & 33.81 & 0/14 \\
\rowcolor[rgb]{0.9,0.9,0.9} \cellcolor[rgb]{1.0,1.0,1.0} & Ko-R1-7B-v2.1 & 40.59 & 28.85 & 38.33 & 39.8 & 35.22 & 33.94 & 26.47 & 29.86 & 42.21 & 29.29 & 48.67 & 46.24 & {\color[HTML]{3166FF} \textbf{64.21}} & 30.67 & 38.70 & 1/14 \\
 & Qwen3-8B & 29.29 & 34.0 & 46.03 & 35.78 & 33.17 & 35.71 & 34.59 & 43.33 & 40.82 & 50.69 & 46.02 & 71.96 & {\color[HTML]{3166FF} \textbf{73.36}} & 59.33 & 46.92 & 1/14 \\
\rowcolor[rgb]{0.9,0.9,0.9} \cellcolor[rgb]{1.0,1.0,1.0} &  Qwen3-8B~{\smaller[2]~(w/ thinking)} & 27.78 & 36.0 & 49.37 & 41.28 & 48.08 & 44.12 & 49.69 & 50.83 & 53.06 & 61.81 & 54.19 & {\color[HTML]{3166FF} \textbf{83.03}} & {\color[HTML]{3166FF} \textbf{76.23}} & {\color[HTML]{3166FF} \textbf{75.33}} & 55.27 & 3/14 \\ \midrule
\textless~20B & Phi-4 (14B) & 33.33 & 34.00 & 41.00 & 36.70 & 37.50 & 37.82 & 42.77 & 44.17 & 43.37 & 45.49 & 41.29 & {\color[HTML]{3166FF} \textbf{72.69}} & 57.38 & 51.33 & 45.32 & 1/14 \\
 & Gemma 3 12B IT & 28.79 & 23.33 & 40.59 & 39.45 & 34.62 & 34.87 & 42.14 & 39.17 & 35.71 & 52.08 & 49.46 & 71.59 & {\color[HTML]{3166FF} \textbf{62.30}} & {\color[HTML]{3166FF} \textbf{68.00}} & 45.82 & 2/14 \\
  & Qwen3-14B & 32.83 & 30.67 & 39.75 & 47.71 & 48.56 & 38.24 & 42.14 & 49.17 & 50.00 & 57.64 & 55.91 & {\color[HTML]{3166FF} \textbf{81.18}} & {\color[HTML]{3166FF} \textbf{75.0}} & {\color[HTML]{3166FF} \textbf{75.33}} & 53.02 & 3/14 \\
 \rowcolor[rgb]{0.9,0.9,0.9} \cellcolor[rgb]{1.0,1.0,1.0} & Qwen3-14B~{\smaller[2]~(w/ thinking)} & 30.81 & 36.67 & 48.95 & 47.71 & 58.65 & 51.26 & 52.2 & 47.50 & 54.59 & 65.28 & 61.94 & {\color[HTML]{3166FF} \textbf{87.82}} & {\color[HTML]{3166FF} \textbf{80.74}} & {\color[HTML]{3166FF} \textbf{82.67}} & 59.48 & 3/14 \\\midrule
\textless~32B & Aya Expanse 32B & 24.75 & 18.67 & 24.27 & 32.11 & 20.67 & 21.01 & 27.67 & 34.17 & 23.98 & 32.64 & 31.83 & 53.14 & 45.49 & 38.67 & 31.26 & 0/14 \\
 & Mistral Small 3.1 Instruct (24B) & 27.27 & 33.33 & 43.51 & 39.45 & 39.90 & 38.66 & 42.77 & 47.50 & 41.33 & 52.08 & 49.89 & {\color[HTML]{3166FF} \textbf{79.34}} & {\color[HTML]{3166FF} \textbf{68.03}} & {\color[HTML]{3166FF} \textbf{72.00}} & 49.49 & 3/14 \\
& Gemma 3 27B IT & 29.80 & 31.33 & 44.35 & 33.94 & 38.46 & 43.28 & 40.88 & 43.33 & 44.39 & 52.08 & 58.49 & 81.18 & {\color[HTML]{3166FF} \textbf{73.36}} & {\color[HTML]{3166FF} \textbf{72.67}} & 51.03 & 2/14 \\
 & Qwen3-30B-A3B & 31.31 & 26.00 & 47.28 & 35.78 & 45.19 & 35.71 & 44.03 & 49.17 & 47.96 & 56.60 & 53.98 & {\color[HTML]{3166FF} \textbf{83.03}} & {\color[HTML]{3166FF} \textbf{77.87}} & {\color[HTML]{3166FF} \textbf{72.00}} & 52.33 & 3/14 \\
 & Qwen3-32B & 33.33 & 35.33 & 53.14 & 43.12 & 57.69 & 45.80 & 53.46 & 53.33 & 48.98 & 61.46 & 60.86 & {\color[HTML]{3166FF} \textbf{85.24}} & {\color[HTML]{3166FF} \textbf{84.84}} & {\color[HTML]{3166FF} \textbf{84.00}} & 58.86 & 3/14 \\
 & EXAONE 4.0 32B & 39.90 & 43.33 & 57.32 & 50.46 & 54.81 & 53.36 & 49.69 & 57.50 & 55.61 & 60.07 & 58.49 & {\color[HTML]{3166FF} \textbf{83.76}} & {\color[HTML]{3166FF} \textbf{77.05}} & {\color[HTML]{3166FF} \textbf{84.00}} & 60.01 & 3/14 \\ 
\rowcolor[rgb]{0.9,0.9,0.9} \cellcolor[rgb]{1.0,1.0,1.0} & Qwen3-30B-A3B~{\smaller[2]~(w/ thinking)} & 35.35 & 37.33 & 50.21 & 45.87 & 57.21 & 50.0 & 48.43 & 54.17 & 54.08 & 70.49 & 59.78 & {\color[HTML]{3166FF} \textbf{88.19}} & {\color[HTML]{3166FF} \textbf{84.43}} & {\color[HTML]{3166FF} \textbf{84.67}} & 60.52 & 3/14 \\
\rowcolor[rgb]{0.9,0.9,0.9} \cellcolor[rgb]{1.0,1.0,1.0} & Qwen3-32B~{\smaller[2]~(w/ thinking)} & 33.33 & 34.67 & 55.23 & 43.12 & 55.29 & 47.06 & 54.09 & 57.50 & 55.61 & 70.14 & {\color[HTML]{3166FF} \textbf{66.88}} & 87.08 & {\color[HTML]{3166FF} \textbf{81.56}} & {\color[HTML]{3166FF} \textbf{88.67}} & 61.14 & 3/14 \\
\rowcolor[rgb]{0.9,0.9,0.9} \cellcolor[rgb]{1.0,1.0,1.0} & QwQ 32B & 35.35 & 39.33 & 55.65 & 44.04 & 60.58 & 55.04 & 57.23 & 56.67 & {\color[HTML]{3166FF} \textbf{64.29}} & 71.53 & {\color[HTML]{3166FF} \textbf{66.24}} & {\color[HTML]{3166FF} \textbf{88.93}} & {\color[HTML]{3166FF} \textbf{84.43}} & {\color[HTML]{3166FF} \textbf{88.67}} & 63.94 & 5/14 \\ 
   \rowcolor[rgb]{0.9,0.9,0.9} \cellcolor[rgb]{1.0,1.0,1.0} & EXAONE 4.0 32B~{\smaller[2]~(w/ thinking)} & 41.92 & 39.33 & 60.25 & 59.63 & 65.38 & 57.56 & {\color[HTML]{3166FF} \textbf{66.67}} & {\color[HTML]{3166FF} \textbf{64.17}} & {\color[HTML]{3166FF} \textbf{69.90}} & 75.69 & {\color[HTML]{3166FF} \textbf{68.60}} & 88.56 & {\color[HTML]{3166FF} \textbf{83.61}} & {\color[HTML]{3166FF} \textbf{87.33}} & 67.67 & 6/14 \\ \midrule
\multirow{6}{*}{\textgreater~70B} & Llama 3.3 70B Instruct & 32.83 & 41.33 & 46.44 & 42.20 & 42.31 & 38.66 & 49.69 & 54.17 & 43.37 & 51.39 & 56.77 & {\color[HTML]{3166FF} \textbf{85.98}} & {\color[HTML]{3166FF} \textbf{71.72}} & {\color[HTML]{3166FF} \textbf{74.00}} & 53.24 & 3/14 \\
 & C4AI Command A (111B) & 41.92 & 34.00 & 51.46 & 49.54 & 45.67 & 45.80 & 46.54 & 54.17 & 52.04 & 61.46 & 57.63 & {\color[HTML]{3166FF} \textbf{83.39}} & {\color[HTML]{3166FF} \textbf{79.10}} & {\color[HTML]{3166FF} \textbf{80.67}} & 57.48 & 3/14 \\
  & Llama-4-Scout-17B-16E-Instruct & 35.86 & 34.00 & 53.14 & 50.46 & 47.12 & 43.70 & 47.80 & 54.17 & 47.45 & 61.46 & {\color[HTML]{3166FF} \textbf{68.17}} & {\color[HTML]{3166FF} \textbf{81.92}} & {\color[HTML]{3166FF} \textbf{85.25}} & {\color[HTML]{3166FF} \textbf{82.67}} & 58.14 & 4/14 \\
 & DeepSeek V3 (671B) & 38.89 & 36.00 & 56.49 & 48.62 & 50.00 & 42.44 & 47.80 & 52.50 & 53.06 & {\color[HTML]{3166FF} \textbf{71.53}} & 64.95 & {\color[HTML]{3166FF} \textbf{87.82}} & {\color[HTML]{3166FF} \textbf{87.70}} & {\color[HTML]{3166FF} \textbf{84.67}} & 60.77 & 4/14 \\
 & Qwen3-235B-A22B & 38.38 & 39.33 & 54.81 & 45.87 & 55.77 & 49.58 & 52.83 & {\color[HTML]{3166FF} \textbf{60.0}} & 57.65 & 64.24 & {\color[HTML]{3166FF} \textbf{70.97}} & 86.72 & {\color[HTML]{3166FF} \textbf{86.07}} & {\color[HTML]{3166FF} \textbf{84.67}} & 62.12 & 4/14 \\
  & Llama-4-Maverick-17B-128E-Instruct & 38.38 & 41.33 & 62.34 & 55.96 & 61.06 & 56.72 & 56.60 & {\color[HTML]{3166FF} \textbf{69.17}} & 66.84 & 75.00 & {\color[HTML]{3166FF} \textbf{76.34}} & 89.67 & {\color[HTML]{3166FF} \textbf{90.98}} & {\color[HTML]{3166FF} \textbf{90.67}} & 68.10 & 4/14 \\
\rowcolor[rgb]{0.9,0.9,0.9} \cellcolor[rgb]{1.0,1.0,1.0} & Qwen3-235B-A22B~{\smaller[2]~(w/ thinking)} & 43.43 & 42.67 & 56.90 & 57.80 & 69.23 & 61.76 & 61.64 & 59.17 & {\color[HTML]{3166FF} \textbf{62.76}} & {\color[HTML]{3166FF} \textbf{71.88}} & {\color[HTML]{3166FF} \textbf{77.42}} & {\color[HTML]{3166FF} \textbf{89.67}} & {\color[HTML]{3166FF} \textbf{89.34}} & {\color[HTML]{3166FF} \textbf{88.00}} & 68.22 & 6/14 \\
\rowcolor[rgb]{0.9,0.9,0.9} \cellcolor[rgb]{1.0,1.0,1.0} & DeepSeek R1 (671B) & 45.45 & 38.00 & 61.51 & 57.80 & 66.83 & 60.50 & {\color[HTML]{3166FF} \textbf{69.18}} & {\color[HTML]{3166FF} \textbf{68.33}} & {\color[HTML]{3166FF} \textbf{64.80}} & 79.51 & {\color[HTML]{3166FF} \textbf{81.29}} & {\color[HTML]{3166FF} \textbf{92.99}} & {\color[HTML]{3166FF} \textbf{94.67}} & {\color[HTML]{3166FF} \textbf{92.67}} & 71.33 & 7/14 \\ \midrule
\multicolumn{17}{l}{Closed Models} \\ \midrule
 & GPT-4.1 mini~{\smaller[2]~(2024-04-14)} & 38.38 & 32.00 & 56.90 & 56.88 & 54.81 & 48.32 & 50.31 & 50.00 & 55.10 & 67.36 & {\color[HTML]{3166FF} \textbf{72.47}} & {\color[HTML]{3166FF} \textbf{90.77}} & {\color[HTML]{3166FF} \textbf{81.97}} & {\color[HTML]{3166FF} \textbf{90.00}} & 62.18 & 4/14 \\
\rowcolor[rgb]{0.9,0.9,0.9} \cellcolor[rgb]{1.0,1.0,1.0} & o3-mini~{\smaller[2]~(2025-01-31)} & 38.38 & 38.67 & 51.46 & 47.71 & 60.10 & 47.06 & 50.31 & 52.50 & 57.65 & 64.24 & {\color[HTML]{3166FF} \textbf{76.56}} & 88.93 & {\color[HTML]{3166FF} \textbf{80.33}} & {\color[HTML]{3166FF} \textbf{91.33}} & 62.05 & 3/14 \\
\rowcolor[rgb]{0.9,0.9,0.9} \cellcolor[rgb]{1.0,1.0,1.0} & Grok-3-mini-beta & 37.37 & 34.67 & 57.74 & 48.62 & 67.79 & 49.58 & {\color[HTML]{3166FF} \textbf{65.41}} & 54.17 & 64.80 & 66.32 & {\color[HTML]{3166FF} \textbf{73.98}} & {\color[HTML]{3166FF} \textbf{91.14}} & {\color[HTML]{3166FF} \textbf{84.84}} & {\color[HTML]{3166FF} \textbf{90.0}} & 65.08 & 5/14 \\
 & Grok-3-beta & 42.42 & 41.33 & 59.00 & 55.96 & 63.94 & 57.98 & {\color[HTML]{3166FF} \textbf{62.89}} & {\color[HTML]{3166FF} \textbf{61.67}} & {\color[HTML]{3166FF} \textbf{66.84}} & 70.49 & {\color[HTML]{3166FF} \textbf{77.85}} & {\color[HTML]{3166FF} \textbf{89.30}} & {\color[HTML]{3166FF} \textbf{91.80}} & {\color[HTML]{3166FF} \textbf{94.67}} & 68.37 & 7/14 \\
\rowcolor[rgb]{0.9,0.9,0.9} \cellcolor[rgb]{1.0,1.0,1.0} & o4-mini~{\smaller[2]~(2025-04-16)} & 37.37 & 46.0 & 62.34 & 49.54 & 69.23 & 55.04 & {\color[HTML]{3166FF} \textbf{66.67}} & 59.17 & {\color[HTML]{3166FF} \textbf{67.35}} & 76.74 & {\color[HTML]{3166FF} \textbf{82.15}} & {\color[HTML]{3166FF} \textbf{93.36}} & {\color[HTML]{3166FF} \textbf{89.75}} & {\color[HTML]{3166FF} \textbf{92.0}} & 69.65 & 6/14 \\
 & GPT-4.1 & 47.47 & 50.00 & {\color[HTML]{3166FF} \textbf{66.95}} & {\color[HTML]{3166FF} \textbf{63.30}} & 66.83 & {\color[HTML]{3166FF} \textbf{59.24}} & {\color[HTML]{3166FF} \textbf{67.30}} & {\color[HTML]{3166FF} \textbf{66.67}} & {\color[HTML]{3166FF} \textbf{68.37}} & 80.21 & {\color[HTML]{3166FF} \textbf{82.80}} & {\color[HTML]{3166FF} \textbf{94.10}} & {\color[HTML]{3166FF} \textbf{92.62}} & {\color[HTML]{3166FF} \textbf{94.67}} & 72.99 & 10/14 \\
\rowcolor[rgb]{0.9,0.9,0.9} \cellcolor[rgb]{1.0,1.0,1.0} & o3 & 45.96 & 41.33 & 71.13 & 57.80 & 73.56 & {\color[HTML]{3166FF} \textbf{61.76}} & {\color[HTML]{3166FF} \textbf{70.44}} & {\color[HTML]{3166FF} \textbf{65.83}} & {\color[HTML]{3166FF} \textbf{71.94}} & {\color[HTML]{3166FF} \textbf{77.43}} & {\color[HTML]{3166FF} \textbf{87.96}} & {\color[HTML]{3166FF} \textbf{92.99}} & {\color[HTML]{3166FF} \textbf{91.39}} & {\color[HTML]{3166FF} \textbf{94.67}} & 73.60 & 9/14 \\
 & Claude 3.7 Sonnet & 55.56 & 53.33 & {\color[HTML]{3166FF} \textbf{73.22}} & {\color[HTML]{3166FF} \textbf{63.30}} & 68.27 & {\color[HTML]{3166FF} \textbf{64.29}} & {\color[HTML]{3166FF} \textbf{67.92}} & {\color[HTML]{3166FF} \textbf{77.50}} & {\color[HTML]{3166FF} \textbf{73.47}} & 70.83 & {\color[HTML]{3166FF} \textbf{81.51}} & {\color[HTML]{3166FF} \textbf{94.46}} & {\color[HTML]{3166FF} \textbf{92.21}} & {\color[HTML]{3166FF} \textbf{93.33}} & 74.52 & 10/14 \\
\rowcolor[rgb]{0.9,0.9,0.9} \cellcolor[rgb]{1.0,1.0,1.0} & Claude 3.7 Sonnet~{\smaller[2]~(w/ thinking)} & 50.00 & {\color[HTML]{3166FF} \textbf{56.00}} & {\color[HTML]{3166FF} \textbf{77.41}} & {\color[HTML]{3166FF} \textbf{74.31}} & 78.85 & {\color[HTML]{3166FF} \textbf{70.17}} & {\color[HTML]{3166FF} \textbf{75.47}} & {\color[HTML]{3166FF} \textbf{70.00}} & {\color[HTML]{3166FF} \textbf{81.12}} & {\color[HTML]{3166FF} \textbf{76.39}} & {\color[HTML]{3166FF} \textbf{84.09}} & {\color[HTML]{3166FF} \textbf{93.36}} & {\color[HTML]{3166FF} \textbf{93.03}} & {\color[HTML]{3166FF} \textbf{92.67}} & 77.70 & 12/14 \\
\rowcolor[rgb]{0.9,0.9,0.9} \cellcolor[rgb]{1.0,1.0,1.0} & o1 & 54.55 & 49.33 & {\color[HTML]{3166FF} \textbf{71.55}} & {\color[HTML]{3166FF} \textbf{65.14}} & 75.48 & 67.23 & {\color[HTML]{3166FF} \textbf{76.73}} & {\color[HTML]{3166FF} \textbf{78.33}} & {\color[HTML]{3166FF} \textbf{78.06}} & {\color[HTML]{3166FF} \textbf{83.33}} & {\color[HTML]{3166FF} \textbf{88.39}} & {\color[HTML]{3166FF} \textbf{94.10}} & {\color[HTML]{3166FF} \textbf{95.49}} & {\color[HTML]{3166FF} \textbf{96.67}} & 78.09 & 10/14 \\ \bottomrule
\end{tabular}
}
\caption{The break down results for all KNPLs in \pro. \colorbox{gray!20}{The gray-shaded models} are the dense-reasoning models. The {\color[HTML]{3166FF} \textbf{blue}} scores indicate that the LLM obtain the license. The details for the pass criteria of each license are described in Appendix~\ref{appendix:pro_scoring}.}
\label{tab:pro_category_result_full}
\end{table*}

%% file: resources/tab_reasoning_vs_nonreasoning.tex
\begin{table*}[!t]
\centering
\footnotesize
\resizebox{\textwidth}{!}{
    \begin{tabular}{@{}r|cccc|ccc|cc|ccccc@{}}
    \toprule
     \multicolumn{1}{l|}{Domain} & \multicolumn{4}{c|}{Law} & \multicolumn{3}{c|}{Tax \& Accounting} & \multicolumn{2}{c|}{Value Estimation} & \multicolumn{5}{c|}{Medical} \\ \midrule
     \multicolumn{1}{l|}{Names of KNPLs}   & \rotatebox{80}{Judicial Scrivener} & \rotatebox{80}{Lawyer} & \rotatebox{80}{Public Labor Attorney} & \rotatebox{80}{Patent Attorney} & \rotatebox{80}{Public Accountant (CPA)}  & \rotatebox{80}{Tax Accountant} & \rotatebox{80}{Customs Broker} & \rotatebox{80}{Damage Adjuster (CDA)} & \rotatebox{80}{Appraiser} & \rotatebox{80}{Doctor of Korean Medicine} & \rotatebox{80}{Dentist} & \rotatebox{80}{Pharmacist} & \rotatebox{80}{Herb Pharmacist} & \rotatebox{80}{Physician} \\ \midrule
        EXAONE 4.0 1.2B & 7.73 & 18.18 & 10.48 & 46.40$^{**}$ & 26.55$^*$ & -12.14 & 32.72$^{**}$ & 20.74$^*$ & 17.83 & 2.18 & 4.57 & 17.02$^{**}$ & 12.75 & 13.43 \\ \hline
        Qwen3-1.7B & 31.58$^*$ & 22.58 & 2.70 & 8.57 & 17.78 & 22.00 & 23.68 & 13.89 & 24.56$^*$ & 46.59$^{**}$ & 25.53$^{**}$ & 33.33$^{**}$ & 46.59$^{**}$ & 34.00$^{**}$ \\ \hline
        Qwen3-8B & -5.17 & 5.88 & 7.27 & 15.38 & 44.93$^{**}$ & 23.53$^{**}$ & 43.64$^{**}$ & 17.31 & 30.00$^{**}$ & 21.92$^{**}$ & 17.76$^{**}$ & 15.38$^{**}$ & 21.92$^{**}$ & 26.97$^{**}$ \\ \hline
        Qwen3-14B & 7.69 & 10.87 & 23.16$^{**}$ & 3.85 & 21.78$^{**}$ & 25.27$^{**}$ & 31.34$^{**}$ & 0.00 & 10.20 & 15.66$^{**}$ & 15.00$^{**}$ & 7.73$^{**}$ & 15.66$^{**}$ & 7.96 \\ \hline
        Qwen3-30B-A3B & 12.90 & 43.59$^{**}$ & 6.19 & 28.21$^*$ & 26.60$^{**}$ & 40.00$^{**}$ & 10.00 & 10.17 & 12.77 & 24.54$^{**}$ & 10.76 & 6.22$^*$ & 24.54$^{**}$ & 17.59$^{**}$ \\ \hline
        Qwen3-32B & 0.00 & -1.89 & 3.94 & 0.00 & -4.17 & 2.75 & 1.18 & 7.81 & 13.54 & 14.12 $^{**}$ & 9.89$^*$ & 2.16 & 14.12$^{**}$ & 5.56 \\ \hline
        EXAONE 4.0 32B & 5.06 & -9.23 & 5.11 & 18.17 & 19.28$^{**}$ & 7.87 & 34.17$^{**}$ & 11.60 & 25.70$^{**}$ & 26.00$^{**}$ & 17.29$^{**}$ & 5.73$^*$ & 8.51$^{**}$ & 3.96 \\ \hline
        Qwen3-235B-A22B & 13.16 & 8.47 & 3.82 & 26.00$^*$ & 24.14$^{**}$ & 24.58$^{**}$ & 16.67$^*$ & -1.39 & 8.85 & 11.89$^{**}$ & 9.09$^*$ & 3.40 & 11.89$^*$ & 3.94 \\ \hline
        Llama 3.3 70B \textsc{vs} R1-Distill-Llama & 7.68 & -8.06 & -54.78$^{**}$ & 19.57 & 45.45$^{**}$ & 11.95 & 12.64 & -1.55 & 17.64$^*$ & 12.84$^*$ & 16.29$^{**}$ & 3.86 & 6.29 & 16.22$^{**}$ \\ \hline
        DeepSeek V3 \textsc{vs} DeepSeek R1 & 16.87 & 5.56 & 8.89 & 18.88 & 33.66$^{**}$ & 42.55$^{**}$ & 44.73$^{**}$ & 30.15$^{**}$ & 22.13$^{**}$ & 11.16$^{**}$ & 25.16$^{**}$ & 5.89$^{**}$ & 7.95$^{**}$ & 9.45$^{**}$ \\ \hline
        Claude 3.7 Sonnet & -10.01 & 5.01 & 5.72 & 17.39$^*$ & 15.50$^{**}$ & 9.15 & 11.12$^*$ & -9.68 & 10.41$^*$ & 7.85$^*$ & 3.17 & -1.16 & 0.89 & -0.71 \\ \hline
    \end{tabular}
}
\caption{The performance gains (\%) when the reasoning is enabled for each model pair. Llama 3.3 70B stands for Llama 3.3 70B Instruct model and R1-Distill-Llama for DeepSeek-R1-Distill-Llama-70B~\citep{deepseekai2025deepseekr1incentivizingreasoningcapability}. $^*$ and $^{**}$ denote the statistical significance by Two-proportion Z-test, indicating p-value < 0.05 and < 0.01, respectively.}
\label{tab:reasoning_vs_non_reasoning}
\end{table*}

%% file: resources/tab_redux_year_test_info.tex
\scriptsize
\begin{table*}
  \centering
    \fontsize{8}{9.5}\selectfont

  \begin{tabularx}{0.95\textwidth}{@{}p{1.5cm}X@{}}
    \toprule
    \textbf{Year} & \textbf{Tests} \\
    \midrule
    2005 & Master Craftsman Construction Equipment Maintenance, Master Craftsman Building General Work, Master Craftsman Precious Metal Processing, Master Craftsman Confectionary Making, Master Craftsman Casting, Master Craftsman Sheet-Metal \& Boiler Making \\ \midrule
    2008 & Master Craftsman Architectural Carpentering, Master Craftsman Surface Treatment \\ \midrule
    2010 & Master Craftsman Railway Vehicles Maintenance \\ \midrule
    2014 & Master Craftsman Welding \\ \midrule
    2016 & Master Craftsman Steel Making \\ \midrule
    2017 & Master Craftsman Metal Material, Master Craftman Metal Mould, Master Craftsman Cook \\ \midrule
    2018 & Master Craftsman Gas, Master Craftsman Machinery Maintenance, Master Craftsman Plumbing, Master Craftsman Rolling, Master Craftsman Energy Management, Master Craftsman Hazardous Material, Master Craftsman Motor Vehicles Maintenance, Master Craftsman Electricity, Master Craftman Electronics, Master Craftsman Iron Making \\ \midrule
    2020 & Engineer Radio Electronic Communication, Engineer Floral Design \\ \midrule
    2021 & Engineer Construction Equipment, Engineer Machinery Design, Engineer Agricultural Health and Safety, Engineer Leak Nondestructive Testing, Engineer Radiation Nondestructive Testing, Engineer Biology Classification—Animal, Engineer Aquaculture, Engineer Visual Communication Design, Engineer Eddy Current Nondestructive Testing, Engineer Welding, Engineer Biomedical, Engineer Magnetic Nondestructive Testing, Engineer Electric Railway, Engineer Computer, Engineer Concrete, Engineer Explosives Handling \\ \midrule
    2022 & Engineer Gas, Engineer Construction Safety, Engineer Construction Material Testing, Engineer Architecture, Engineer Building Facilities, Engineer Air-Conditioning Refrigerating Machinery, Engineer Transportation, Engineer Metal, Engineer Meteorology, Engineer Air Pollution Environmental, Engineer Urban Planning, Engineer Bioprocess, Engineer Forest, Engineer Industrial Safety, Engineer Industrial Hygiene Management, Engineer Plant Maintenance, Engineer Fire Protection System—Mechanical, Engineer Fire Protection System—Electrical, Engineer Noise \& Vibration, Engineer Water Pollution Environmental, Engineer Elevator, Engineer Plant Protection, Engineer Food Processing Safety, New and Renewable Energy Equipment (Photovoltaic) Engineer, Engineer Interior Architecture, Engineer Energy Management, Engineer Greenhouse Gas Management, Engineer Organic Agriculture, Engineer Ergonomics, Engineer General Machinery, Engineer Motor Vehicles Maintenance, Engineer in Nature Environment and Ecological Restoration, Engineer Electric Work, Engineer Electricity, Engineer Computer System Application, Engineer Electronics, Engineer Information Processing, Engineer Landscape Architecture, Engineer Seeds, Engineer Cadastral Surveying, Engineer Railroad Signal Apparatus, Engineer Ultrasonic Nondestructive Testing, Engineer Livestock, Engineer Surveying Geo-Spatial Information, Engineer Penetrante Nondestructive Testing, Engineer Colorist, Engineer Civil Engineering, Engineer Soil Environment, Master Craftsman Telecommunication Apparatus, Engineer Wastes Treatment, Engineer Quality Management, Engineer Ocean Environment, Engineer Chemical Industry, Fire Investigation \& Evaluation Engineer, Engineer Chemical Analysis \\ \midrule
    2023 & Engineer Radio Telecommunication Equipment, Engineer Broadcasting Communication, Engineer Information Communication \\
    \bottomrule
  \end{tabularx}
    \caption{Redux Years and National Qualification Test Additions}
  \label{tab:redux_detail_years_NTQ}
\end{table*}

%% file: resources/tab_kmmlu_error_ex.tex
\begin{table*}[!ht]
\centering
\resizebox{\textwidth}{!}{
\begin{tabular}{ll}\toprule
\textbf{Error Type} & 
\textbf{Examples} \\ \midrule
Ill-posed Question & \begin{tabular}[c]{@{}l@{}} Category: Political science and sociology\\ A, B에 대한 설명으로 옳은 것만을 \textless{}보기\textgreater{}에서 고르면? \\ \textit{\textcolor{gray}{What is the correct explanation about A, B in \textless{}Reference\textgreater{}?}}\end{tabular} \\ \midrule
Leaked Answer & \begin{tabular}[c]{@{}l@{}}Category: Ecology\\ 산복수로에서 쌓기공작물의 높이가 3m이고, 수로깊이가 1m일 때 수로받이의 \\ 근사적 길이는? (문제 오류로 현재 복원중입니다. 보기 내용을 아시는 분들께서는 \\ 오류 신고를 통하여 보기 작성 부탁 드립니다. 정답은 3번입니다.) \\ \textit{\textcolor{gray}{What is the approximate length of the culvert if the pile is 3 meters high and the}} \\ \textit{\textcolor{gray}{ channel is 1 meter deep? (This is currently being restored due to a question error.}} \\ \textit{\textcolor{gray}{If you know the referebce, please report the error. The correct answer is 3.)}} \end{tabular} \\ \midrule
Notation Error & \begin{tabular}[c]{@{}l@{}} Category: Math \\ 다항식 x2017-1을 x2-x로 나누었을 때의 나머지를 R(x)라 할 때, R(2017)의 값은?\\ \textit{\textcolor{gray}{If the remainder of the polynomial x2017-1 divided by x2-x is called R(x), what is}} \\ \textit{\textcolor{gray}{the value of R(2017)?}}\end{tabular} \\ \midrule
Bad Clarity & \begin{tabular}[c]{@{}l@{}} Category: Education \\ 정신분석 상담과 행동주의 상담의 공통점에 해당하는 것은?\\ A. 상담과정에서 과거 경험보다 미래 경험을 중시한 다 .\\ B. 상 담 기 법 보 다 는 상 담 자 의 인 간 적 자 질 과 진 솔 한 태 도 를 중 시 한 다 .\\ C. 인간의 행동을 인과적 관계로 해석하는 결정론적 관점을 가진 다 .\\ D. 비합리적 신념을 인식하고 수정하는 논박 과정을 중시한 다 . \\ \textit{\textcolor{gray}{Which is a common feature of psychoanalytic counseling and behavioral counseling?}} \\
\textit{\textcolor{gray}{A. Emphasizes future experiences over past experiences in the counseling process.}} \\
\textit{\textcolor{gray}{B. Prioritizes the counselor’s human qualities and sincerity over counseling techniques.}} \\
\textit{\textcolor{gray}{C. Interprets human behavior through a deterministic perspective based on causal relationships.}} \\
\textit{\textcolor{gray}{D. Focuses on the disputation process to recognize and modify irrational beliefs.}}
\end{tabular} \\ \bottomrule
\end{tabular}}
\caption{Examples of error types in KMMLU. Each example demonstrates a specific issue that impacts the reliability of the benchmark. \textcolor{gray}{Gray} text represents translation of the examples in English}
\label{tab:example_error_types}
\end{table*}